\documentclass{article}

\PassOptionsToPackage{numbers,compress}{natbib}

\usepackage[accepted]{icml2026}

\PassOptionsToPackage{dvipsnames}{xcolor}
\definecolor{NavyBlue}{RGB}{0,0,128}
\usepackage[allcolors=NavyBlue,colorlinks=true]{hyperref}
\usepackage[T1]{fontenc}
\usepackage[utf8]{inputenc}
\usepackage{url}
\usepackage{microtype}
\usepackage{booktabs}
\usepackage{nicefrac}
\usepackage{xspace}
\usepackage{amsmath,amssymb,amsthm,mathtools}
\usepackage{multirow,makecell,threeparttable,diagbox,adjustbox}
\usepackage{wrapfig}
\usepackage{caption}
\usepackage{subcaption}
\usepackage{natbib}
\usepackage{algorithm}
\usepackage[noend]{algpseudocode}

\usepackage{comment}
\usepackage{prettyref}
\usepackage{multicol}
\usepackage[textsize=scriptsize]{todonotes}
\usepackage{tabularx}

\definecolor{colorcomment}{RGB}{160, 190, 210}%
\makeatletter
\algnewcommand{\LineComment}[1]{\Statex \hskip\ALG@thistlm \(\triangleright\) 
{\color{colorcomment}#1}}
\makeatother

\makeatletter
\algnewcommand{\IndentLineComment}[1]{\Statex \hskip\ALG@tlm \(\triangleright\) {\color{colorcomment}#1}}
\makeatother

\newcommand\policy{\ensuremath{\pi}}

\newcommand\state{s}

\newcommand\reward{r}

\newcommand\QFunc{\ensuremath{\ensuremath{Q}}}

\newcommand\Num{\ensuremath{N}}

\newcommand\action{\ensuremath{a}}

\newcommand\actionSpace{\ensuremath{\mathcal{A}}}

\newcommand\horizon{\ensuremath{H}}

\newcommand\PHUCT{\ensuremath{\text{P}\mathcal{H}\text{-UCT}}} 

\newcommand\PHUCBME{\ensuremath{\text{P}\mathcal{H}\text{-UCB}\text{-ME}}}
\newcommand\PHUCTME{\ensuremath{\text{P}\mathcal{H}\text{-UCT}\text{-ME}}} 

\newcommand\PH{\ensuremath{\text{P}\mathcal{H}}}
\newcommand{\Uent}{\ensuremath{\mathcal{U}_{\text{ent}}}}
\newcommand{\Udiv}{\ensuremath{\mathcal{U}_{\text{div}}}}
\newcommand{\TopK}{\operatorname{Top\text{-}K}}
\newcommand{\HH}{\mathcal{H}}

\def \argmax {\mathop{\rm arg\,max}}

\newcommand{\algname}{\textsc{MCTD-ME}\xspace}

\newif\iffinal
\finaltrue

\iffinal
    \newcommand{\fix}[1]{#1}
    \newcommand{\XL}[1]{}
    \newcommand{\XLinline}[1]{}
    \newcommand{\SH}[1]{}
    \newcommand{\SHinline}[1]{}
    \newcommand{\CT}[1]{}
    \newcommand{\CTinline}[1]{}
    \newcommand{\DP}[1]{}
    \newcommand{\DPinline}[1]{}
    \newcommand{\note}[1]{}
    \newcommand{\pref}[1]{}
\else
    \newcommand{\fix}[1]{{\color{red} #1}}
    \newcommand{\SH}[1]{\todo[fancyline,color=Maroon!40]{SH: #1}\xspace}
    \newcommand{\DP}[1]{\todo[fancyline,color=Purple!40]{DP: #1}\xspace}
    \newcommand{\XL}[1]{\todo[fancyline,color=NavyBlue!40]{XL: #1}\xspace}
    \newcommand{\XLinline}[1]{\textcolor{ForestGreen}{[XL: #1]}}
    \newcommand{\CT}[1]{\todo[fancyline,color=NavyBlue!40]{CT: #1}\xspace}
    \newcommand{\CTinline}[1]{\textcolor{NavyBlue}{[CT: #1]}}

    \newcommand{\note}[1]{{\color{purple}[XL: #1]}}
    
    \newcommand{\pref}[1]{{\color{blue}(\ref{#1})}}
\fi

\newcommand{\tabref}[1]{Table~\ref{#1}}
\newcommand{\figref}[1]{Fig~\ref{#1}}

\newcommand{\secref}[1]{\S\ref{#1}}
\newcommand{\appref}[1]{Appendix~\ref{#1}}

\newcommand{\algoref}[1]{Algorithm~\ref{#1}}

\newcommand{\paren} [1] {\ensuremath{ \left( {#1} \right) }}

\newcommand{\curlybracket}[1]{\ensuremath{\left\{#1\right\}}}

\theoremstyle{plain}

\theoremstyle{definition}

\theoremstyle{remark}

\begin{document}

\twocolumn[
  \icmltitle{Monte Carlo Tree Diffusion with Multiple Experts for Protein Design}

  \icmlsetsymbol{equal}{*}

  \begin{icmlauthorlist}
    \icmlauthor{Xuefeng Liu}{equal,uchicago}
    \icmlauthor{Mingxuan Cao}{equal,uchicagods}
    \icmlauthor{Songhao Jiang}{uchicago}
    \icmlauthor{Xiao Luo}{ttic}
    \icmlauthor{Xiaotian Duan}{argonne}
    \icmlauthor{Mengdi Wang}{Princeton}
    \icmlauthor{Tobin R. Sosnick}{uchicagobme}
    \icmlauthor{Jinbo Xu}{ttic}
    \icmlauthor{Rick Stevens}{uchicago,argonne}
  \end{icmlauthorlist}

  \icmlaffiliation{uchicago}{Department of Computer Science, University of Chicago, Chicago, IL, USA}
 \icmlaffiliation{uchicagods}{Data Science Institute, University of Chicago, Chicago, IL, USA} 
  \icmlaffiliation{uchicagobme}{Department of Biochemistry and
Molecular Biology, University of Chicago, Chicago, IL, USA} 
 \icmlaffiliation{ttic}{Toyota Technological Institute at Chicago, Chicago, IL, USA}
  \icmlaffiliation{argonne}{Argonne National Laboratory, Lemont, IL, USA}
  \icmlaffiliation{Princeton}{AI Lab, Princeton University, NJ, USA}

  \icmlcorrespondingauthor{Xuefeng Liu}{Xuefeng@uchicago.edu}
 \icmlcorrespondingauthor{Mingxuan Cao}{caom@uchicago.edu}

  \icmlkeywords{Machine Learning, ICML}

  \vskip 0.3in
]

\printAffiliationsAndNotice{\icmlEqualContribution}

\begin{abstract}

The goal of protein design is to generate amino acid sequences that fold into functional structures with desired properties. Prior methods combining autoregressive language models with Monte Carlo Tree Search (MCTS) struggle with long-range dependencies and suffer from an impractically large search space. We propose \algname, Monte Carlo Tree Diffusion with Multiple Experts, which integrates masked diffusion models with tree search to enable multi-token planning and efficient exploration {under the guidance of multiple experts}. Unlike autoregressive planners, \algname uses biophysical-fidelity-enhanced diffusion denoising as the rollout engine, jointly revising multiple positions and scaling to large sequence spaces. It further leverages experts of varying capacities to enrich exploration, guided by a pLDDT-based masking schedule that targets low-confidence regions while preserving reliable residues. We propose a novel multi-expert selection rule (\PHUCTME) extends {Shannon-entropy-based UCT to expert ensembles with mutual information.} \algname achieves superior performance on the CAMEO and PDB benchmarks, excelling in protein design tasks such as inverse folding, folding, and conditional design challenges like motif scaffolding on lead optimization tasks.
Our framework is model-agnostic, plug-and-play, and extensible to de novo protein engineering and beyond.

\end{abstract}

\section{Introduction}\XL{shorten}

Large generative models have demonstrated impressive capabilities across domains such as natural language processing \citep{vaswani2017attention} and molecular design \citep{MEYERS20212707}. Yet, conventional decoding strategies—greedy sampling, nucleus sampling, and beam search—remain fundamentally limited in tasks that require strategic planning or optimizing multiple objectives. In molecule or protein design, for example, a generative model must produce sequences that are not only syntactically valid but also meet biochemical, structural, and functional requirements \citep{Liu2024-ERP}. Greedy decoding often gets stuck in suboptimal choices, and beam search—despite its popularity—offers little control over long-range dependencies or diverse trade-offs, especially in long-horizon domains like drug generation or code synthesis \citep{bagal2021molgpt}.

In contrast, planning algorithms such as Monte Carlo Tree Search (MCTS) offer adaptive search through a balance of exploration and exploitation. Since the introduction of UCT \citep{kocsis2006bandit}, MCTS has shown success in domains requiring sequential decision-making, most notably in AlphaGo \citep{silver2017mastering}. Unlike static decoding, MCTS enables dynamic trajectory expansion, backtracking, and selective refinement, making it a promising mechanism to augment generative models with deliberative search. Recent frameworks such as Tree-of-Thoughts \citep{yao2023tree} and RethinkMCTS \citep{Li2024RethinkMCTSRE} illustrate this potential in language and code generation, where search paths can be evaluated and adjusted.

Large language models (LLMs such as GPT \citep{liu2025drugimprovergpt, liu2025scaffoldgpt,liu2025controllablegpt}) have recently been paired with MCTS to improve search in complex generation tasks. For instance, GPT-guided MCTS has enhanced search efficiency in symbolic regression \citep{Li2024DiscoveringMF}, while value-guided MCTS decoding (PPO-MCTS \citep{liu2023dont}) improved the preferability of generated text over standard RLHF outputs. However, GPT-based MCTS approaches face notable pitfalls. Due to autoregressive, token-by-token generation, even a single token error can derail an entire solution, and models often get stuck in suboptimal trajectories because their prior training biases lead to unbalanced exploration-exploitation. In other words, maintaining long-range coherence is challenging and it’s difficult to revise parts of a candidate without regenerating the whole sequence.

Beyond these GPT-specific pitfalls, applying MCTS to biological sequence design raises core problems: (i) {no reliable partial-state evaluator} (i.e., no tractable forward model) to score or prune partial sequences; (ii) {multi-objective constraints} (e.g., structural compatibility, folding efficiency, evolutionary plausibility); (iii) {poor scaling to long sequences} under autoregressive generation, making tree search expensive and prone to exploring unpromising states; and (iv) {limited diversity/biophysical fidelity} when exploration relies on a single pretrained model.
While Diffuser \citep{janner2022diffuser} shows diffusion models can act as planners without explicit forward simulators, diffusion decoding is still largely open-loop, offering limited opportunities for test-time correction.

In this paper, we propose Monte Carlo Tree Diffusion with Multiple Experts (\algname), a novel planning framework that integrates diffusion-based generation with an MCTS search guided by a diverse ensemble of domain experts models. Motivated by the challenges of protein sequence design—where search spaces are vastly larger than small-molecule generation—we leverage diffusion models for their ability to efficiently denoise multiple positions in parallel, and use expert models to narrow the search space by injecting biologically grounded priors. Each node in our tree represents a partially denoised sequence (a partial plan), and expansion corresponds to one step of noise reduction. This structure imbues the diffusion process with MCTS’s adaptive lookahead: evaluating, pruning, and refining paths to focus on promising candidates. 
In addition, we adopt an imitation-style rollout approach, where each expert proposes candidate completions based on the current partial sequence, thereby increasing diversity in the exploration space.
These trajectories guide selection and backpropagation through the tree. Disagreement among experts encourages exploration; consensus promotes exploitation—enabling informed multi-objective planning at test time.

\textbf{Summary of contributions:} 
To our knowledge, this is the first work to incorporate \fix{an ensemble of \emph{experts}} into a diffusion-based planning framework, enabling \fix{multi-expert} Monte Carlo tree search for complex generative problems. By unifying diffusion models with \fix{a multi-critic \emph{evaluation} module and a multi-expert \emph{proposal} module within MCTS}, we introduce \fix{a general inference-time strategy} to steer generation toward goals that single decoders or static samplers struggle with—especially in large, structured domains like protein sequence design, where generation spaces are vast and require both parallel decoding and multi-perspective biophysical-fidelity-aware exploration. 

Second, we propose several technical innovations to realize this integration:
\begin{enumerate}
    \item \fix{A tree-structured diffusion process for partial sequences, where each node represents a masked-infill state and expansion corresponds to a denoising rollout.} We \fix{improve biophysical fidelity via a pLDDT-guided masking policy that focuses edits on structurally uncertain regions}, and we enable \fix{diverse, goal-directed exploration} using a multi-expert rollout mechanism, where each expert proposes candidate completions conditioned on the partial sequence.
    \item 
    \fix{A novel \PHUCTME~selection rule for multi-expert MCTS}, which arbitrates among experts using \fix{cached} uncertainty statistics for stable and efficient search, extending the classic UCT formula with exploration bonuses based on expert disagreement (e.g., score variance or entropy), inspired by entropy-guided UCB~\citep{Liu2024-ERP}.
\end{enumerate}
\vspace{-0.3cm}

Third, we demonstrate empirical gains on long-horizon generation tasks—folding, inverse folding, and motif scaffolding—on CAMEO 2022~\citep{jing2023eigenfold} and a PDB date-split benchmark~\citep{campbell2024generative}. In the lead-optimization regime (equal test-time compute), our planner consistently improves over strong baselines such as ESMFold~\citep{lin2024esmfold_science}, DPLM-2~\citep{wang2024dplm2multimodaldiffusionprotein}, and ProteinMPNN~\citep{dauparas2022proteinmpnn} (we report absolute baseline/final values and $\Delta$). We obtain lower RMSD and higher TMscore in folding, higher AAR and scTM in inverse folding, and improved motif preservation in scaffolding; moreover, we outperform single-expert ablations of RFDiffusion~\citep{Watson2023RFdiffusion}, FoldFlow~\citep{Bose2024FoldFlow}, and ProteinA~\citep{Geffner2025Proteina}, with gains amplifying on longer proteins and more challenging scaffolds where uncertainty is concentrated. \fix{Beyond lead optimization, we additionally evaluate a de novo design setting (all-mask root) in \appref{app:additionalexperiments}.}

\fix{Our results highlight three practical advantages: (i) pLDDT-guided masking concentrates compute on low-confidence regions, improving efficiency and stability; (ii) multi-expert selection leverages complementary pretrained models, yielding robust gains even when individual experts underperform; and (iii) quality improves smoothly with larger search budgets, enabling controllable compute--performance trade-offs.} Importantly, the method is \fix{model-agnostic and modular: it can integrate any set of pretrained experts and black-box critics at test time without retraining}, making it broadly applicable across domains. Overall, this work advances the state-of-the-art in \fix{test-time guided generative planning}, opening up promising directions for applications in drug discovery, program synthesis, and beyond.

\section{Related works}

Recent work shows diffusion models are promising for biomolecular design, including structure-based backbone generation (RFdiffusion \citep{Watson2023DeNovo}), discrete sequence generation with versatile conditioning (DPLM and DPLM-2 \citep{wang2024diffusionlanguagemodelsversatile,wang2024dplm2multimodaldiffusionprotein}), and property-guided atom--bond generation (DiffGui \citep{hu2025diffgui}); however, these methods are typically one-shot and lack test-time refinement. In parallel, tree search has improved generation for code and reasoning (AlphaCode \citep{li2022competition}, RethinkMCTS \citep{Li2024RethinkMCTSRE}, Tree-of-Thoughts \citep{yao2023tree}, PG-TD \citep{zhang2023planning}) and more recently \fix{System-2-style planning for LMs} \fix{\citep{Baek2025MonteCD}}, but largely targets autoregressive GPT-style models rather than diffusion.

Discrete diffusion supports generation via absorbing-state noise (D3PM \citep{Austin2021StructuredDD}) and masked-diffusion LMs simplifying denoising \citep{shi2024simplified, sahoo2024simple}, plus planning (Diffuser \citep{janner2022diffuser}), inverse folding (ProtInvTree \citep{liu2025protinvtreedeliberateproteininverse}), and peptide optimization (PepTune \citep{Tang2024PepTune}). \fix{Closest, masked-diffusion has also been used for planning in diffusion spaces (e.g., path planning) \citep{sahoo2024simple, shi2024simplified}, but these methods plan over trajectories, whereas our setting is protein sequence design with biophysical critics and pLDDT-guided edit control.} While mixture-of-experts strategies have been explored in RL and diffusion-based image generation \citep{fang2024remix,lee2024multi, liu2023active}, they are static at inference. Our approach integrates multiple specialized expert generators within an MCTS-driven diffusion framework for dynamic collaboration, adaptive search, and multi-objective biomolecular design. (Additional details are provided in \appref{app:related}.)

\section{Preliminaries}
\paragraph{Markov decision processes.} 

We formulate generative planning as a finite-horizon MDP $\mathcal{MDP} = \langle \mathcal{S}, \mathcal{A}, H, P, R \rangle$ \citep{puterman2014markov}, where $\mathcal{S}$ is the set of states (e.g., partial or completed sequences), $\mathcal{A}$ is the action space (e.g., denoising steps or token insertions), and $P: \mathcal{S} \times \mathcal{A} \to \mathcal{S}$ defines stochastic transitions. The reward $R: \mathcal{S} \times \mathcal{A} \to \mathbb{R}$ is often defined only on terminal states, and the policy $\pi: \mathcal{S} \to \Delta(\mathcal{A})$ specifies an action distribution (e.g., a diffusion denoiser or autoregressive next-token model). An episode constructs a length-$H$ sequence $y = (y_1, ..., y_H)$. We denote the action-value function $Q^\pi(s, a)$ and value function $V^\pi(s)$.

\paragraph{Diffusion-based generative model.} We review diffusion for sequence generation. Let $y_{0:H}$ denote an $H$-token sequence. Forward noising $q(y_t \mid y_{t-1})$ for $t=1,\ldots,T$ corrupts the sequence until $y_T$ is maximally noisy (e.g., fully masked). A reverse model $p_\phi(y_{t-1} \mid y_t)$ is trained to denoise, enabling sampling by iterating from $y_T \sim p(y_T)$ to $y_0 \sim p_\phi(y_0)$.Crucially, diffusion models naturally support guided generation. When an auxiliary score function $f(y)$ is available—e.g., a property predictor or constraint critic—generation can be biased toward desirable outputs. One classic approach is classifier guidance, where each reverse step is adjusted in the direction of $\nabla_{y_{t-1}} f(y_{t-1})$. Alternatively, reverse distribution can be reweighted as:
$$
p^{\text{guide}}_\phi(y_{t-1} \mid y_t) \propto p_\phi(y_{t-1} \mid y_t) \cdot \exp\left\{ \beta \cdot f(y_{t-1}) \right\},
$$
with $\beta$ controlling guidance strength. We extend this to multiple experts: $K$ evaluators $\{C_k\}_{k=1}^K$ score complete sequences proposed by \fix{ multiple experts $\{\pi^e\}_{e=1}^E$ }.

\vspace{-0.15cm}
\paragraph{MCTS.} Monte Carlo Tree Search (MCTS) balances exploration and exploitation via tree search and simulation. It iterates selection, expansion, evaluation, and backpropagation. Selection uses UCT \citep{kocsis2006bandit}:
\begin{equation}\label{eq:UCB}
\text{UCB}=\QFunc\paren{\state,\action}+c_p\cdot\sqrt{\frac{\log\paren{\Num\paren{\state}}}{\Num\paren{\state,\action}}}, 
\end{equation}
where $Q(s,a)$ is the estimated reward, $N(s)$ and $N(s,a)$ are visit counts, and $c_p$ controls exploration. Expansion adds children, evaluation runs a rollout to obtain a reward, and backpropagation updates $N(\cdot)$ and $Q(\cdot)$. Repetition concentrates search on high-reward regions.

\section{The \algname Algorithm}
\begin{figure*}[t]
  \centering
  \includegraphics[width=0.8\linewidth]{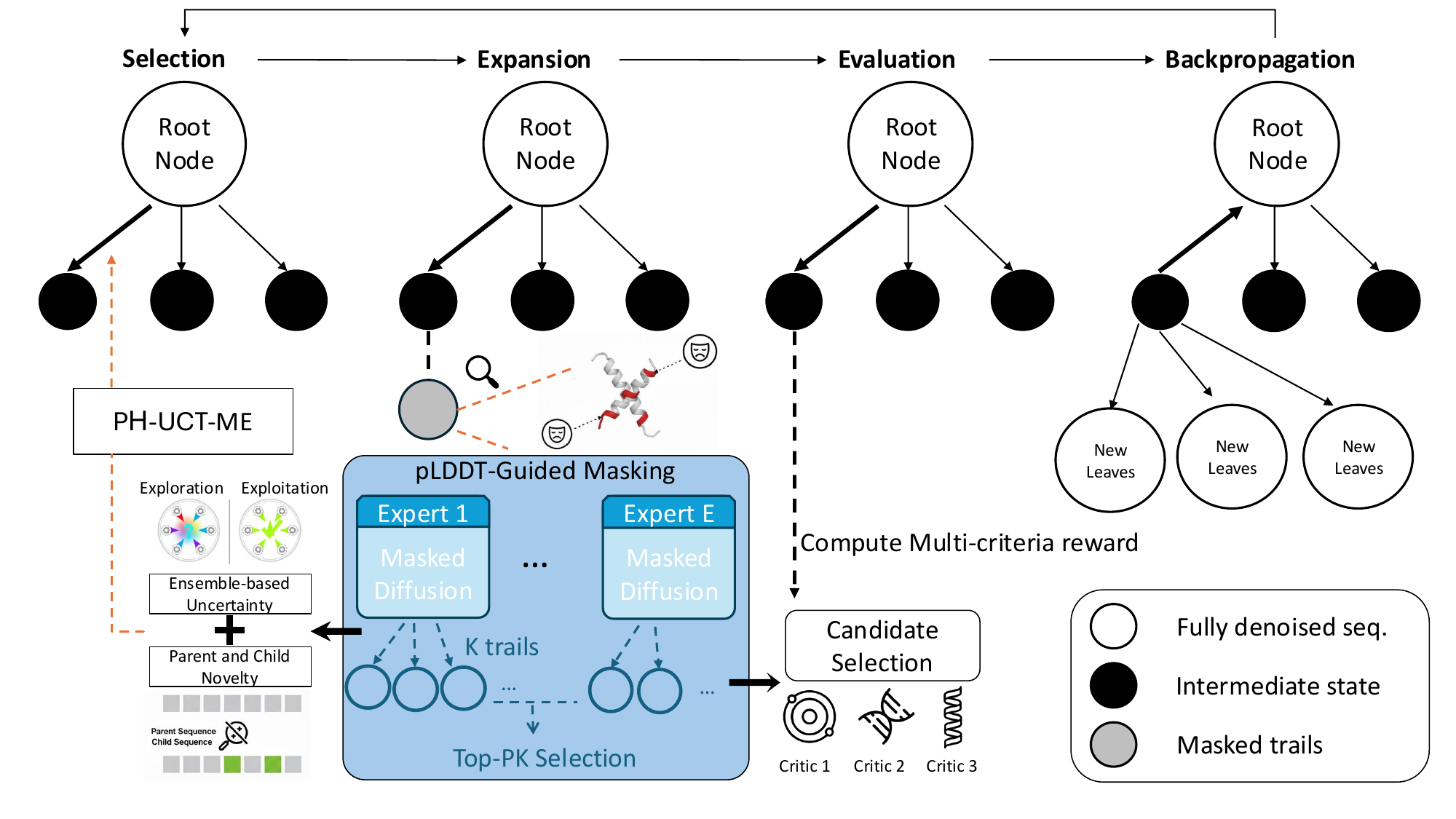}
  \caption{Overview of \algname in diffusion-based protein sequence generation. Nodes are partially denoised sequences expanded via masked diffusion and guided by multi-expert rollouts.
  }
  \label{fig:mctd_overview}
  \vspace{-0.5cm}
\end{figure*}

\begin{figure}[t]
\vspace{-8pt}
\centering
\includegraphics[width=\linewidth]{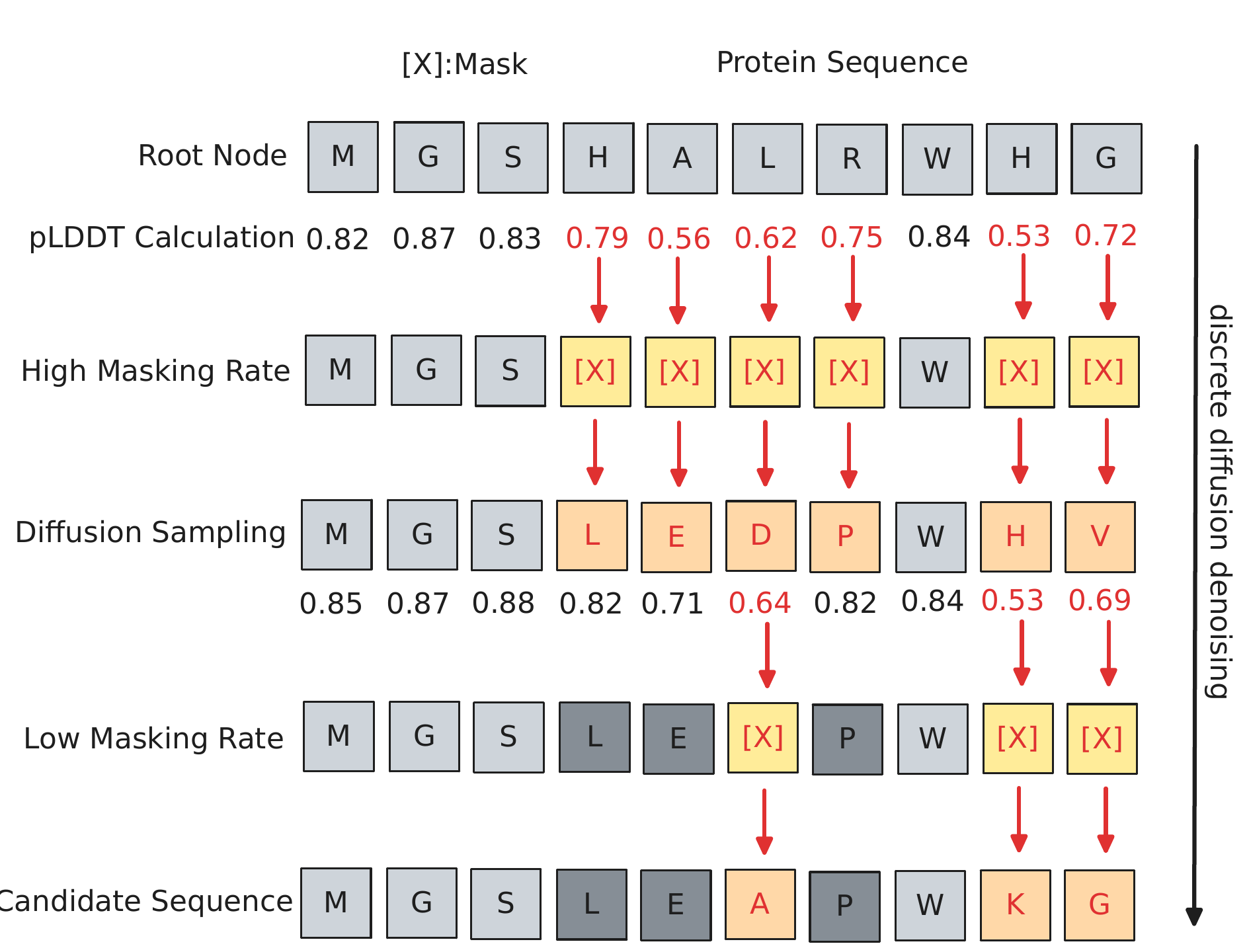}
\caption{Masked discrete diffusion step: low-confidence sites (mask) are resampled while confident tokens remain fixed, yielding a progressively “unmasked” sequence.}
\label{fig:mask-schematic}
\vspace{-10pt}
\end{figure}

We propose Monte Carlo Tree Diffusion with Multi-Experts (\algname), which enhances protein generation by combining masked diffusion with an MCTS planner. 
\fix{An overview of the \algname workflow is shown in \figref{fig:mctd_overview}.}
\algname structures diffusion as a tree, refining partially masked sequences to preserve scaffold integrity while systematically exploring uncertain regions. Search quality and diversity are further improved by incorporating multiple diffusion experts for diverse refinements and by using pLDDT-guided masking to target low-confidence residues, \fix{as illustrated in \figref{fig:mask-schematic}}. These design choices couple the planning strength of tree search with expert-aware diversification, yielding a principled framework for balancing exploration and exploitation in protein modeling. We present the algorithm in \algoref{alg:mctd} (deferred to \appref{app:algorithm}), with each rollout consisting of four key steps as following:

\paragraph{Selection.} 
To navigate the space of protein sequences under the reverse \emph{masked} diffusion framework, we define a tree where each node corresponds to a \emph{complete} sequence (with no masked tokens) at a particular reverse step, and edges represent one reverse transition obtained by applying masked discrete diffusion to a low-confidence subset of tokens. 
{
At each iteration, we employ a novel selection algorithm, {\PHUCTME}—a Policy(P)-guided entropy($\mathcal{H}$)-augmented variant of UCT—designed to identify the most promising path, where ME denotes Multiple Experts.}
Specifically, from a parent node $\state_t$ we choose the transition (i.e., reverse move to $\state_{t-1}$ via a particular mask set) that maximizes the following objective:
\begin{align}
\PHUCTME(\state_t) \;=\; \argmax_{\action\in\actionSpace}\; 
\PHUCBME\!\left((\state_t,\action)\right),\nonumber
\end{align}
with
\begin{align}
\PHUCBME\!& \left((\state_t,\action)\right)
= \QFunc(\state_t,\action)
+ c_p\,\frac{\sqrt{\log \Num(\state_t)}}{1+\Num(\state_t,\action)}  \notag\\
 \;\fix{\pi_{\mathrm{cons},\tau}(\action\mid \state_t)}
& \cdot \paren{\underbrace{w_{\mathrm{ent}}\,\Uent(\state_t,\action)
\;+\; w_{\mathrm{div}}\,\Udiv(\state_t,\action)}_{\text{\PH\text{-ME}~bonus (cached at expansion)}}}.
\end{align}

Here, $\QFunc(\state_t,\action)$ is the current value estimate, $\Num(\cdot)$ are visit counts, $c_p>0$ controls exploration, and $w_{\mathrm{ent}},w_{\mathrm{div}}\!\ge\!0$ weight the predictive-uncertainty and diversity bonuses, respectively. 
In our diffusion setting, the action prior is given by the reverse diffusion model $p_\phi(\state_{t-1}\!\mid\!\state_t)$ over masked edits, but---consistent with our implementation---we \emph{do not} recompute its entropy during selection; instead we cache uncertainty statistics for each expanded child and reuse them during subsequent selections. \fix{We cache a temperature-controlled \emph{consensus prior} 
$\pi_{\mathrm{cons},\tau}(\action\mid \state_t)\!\in\![0,1]$ at expansion—large only when multiple experts assign high mass to the same action—so the UCB exploration is steered toward expert-agreed edits while preserving asymptotics. We define
\begin{align}
\pi_{\mathrm{cons},\tau}(a\mid s)
=\frac{1}{Z_\tau}\exp\!\Big(\tfrac{1}{\tau}\sum_{e=1}^{E}\log p^{(e)}_\phi(a\mid s,\mathcal M)\Big),
\\
Z_\tau=\sum_{a'\in\mathcal A(s)}\exp\!\Big(\tfrac{1}{\tau}\sum_{e=1}^{E}\log p^{(e)}_\phi(a'\mid s,\mathcal M)\Big),
\end{align}
i.e., a temperatured geometric mean across experts.}

\emph{Multi-Expert P$\mathcal{H}$ (P$\mathcal{H}$-ME) bonus.}
Let $\mathcal{M}$ be the mask set chosen at $\state_t$ (e.g., via pLDDT), and let $\{p_\phi^{(e)}\}_{e=1}^{E}$ denote the $E$ expert conditional distributions \fix{(softmax of logits)} used during expansion. 
For the candidate child $\state_{t-1}$ obtained by action $\action$, we define the ensemble-based uncertainty as
\begin{align}
\Uent(\state_t,\action)
= &\HH\!\left(\frac{1}{E}\sum_{e=1}^{E}
p_\phi^{(e)}(\,\cdot\, \mid \state_{t-1},\mathcal{M})\right)\notag \\
& - \frac{1}{E}\sum_{e=1}^{E}\,\HH\!\left(p_\phi^{(e)}(\,\cdot\, \mid \state_{t-1},\mathcal{M})\right).
\label{eq:ensemble-surprisal}
\end{align}
\fix{Here $\HH(p)=-\sum_a p(a)\log p(a)$ is Shannon entropy; the difference equals a BALD/Jensen–Shannon–style disagreement signal, isolating epistemic uncertainty.} 

\fix{We also include a diversity measure, defined as the normalized Hamming distance between parent and child sequences:}
\begin{align}
\Udiv(\state_t,\action)
&= \frac{1}{L}\sum_{i=1}^{L}\,\mathbf{1}\{y^{\text{child}}_i \neq y^{\text{parent}}_i\}.
\end{align}
\fix{
In our multi-model setting, sequences are batched and multi-channel (e.g., amino-acid tokens and structure tokens). 
We evaluate $\Udiv$ over the task-relevant token subset: for \emph{folding}, the difference is computed on structure tokens relative to the root structure; for \emph{inverse folding}, on amino-acid tokens relative to the baseline sequence; and for \emph{motif scaffolding}, on both channels (combined by concatenation or by averaging channel-wise normalized distances).
When batching, we average the per-sequence $\Udiv$ across the batch, thus the result remains in $[0,1]$.
} 
Both $\Uent$ and $\Udiv$ are computed once at expansion and cached for reuse during selection.

An action $\action$ is prioritized if it (i) has high estimated value $Q(\state_t,\action)$, (ii) corresponds to an under-explored branch (UCB exploration), (iii) induces strong ensemble disagreement ($\Uent$ large), indicating information gain, and (iv) introduces sufficient novelty relative to the parent ($\Udiv$ large). 
This encourages the search to balance reward exploitation with uncertainty- and diversity-driven exploration, while the reverse diffusion prior $p_\phi(\state_{t-1}\!\mid\!\state_t)$ maintains consistency with the denoising dynamics.

\noindent\textit{{Remark 1}} (single-expert case).
When $E{=}1$, \eqref{eq:ensemble-surprisal} vanishes (\,$\Uent{=}0$\,). In \textsc{MCTD-1}, we therefore use the predictive (Shannon) entropy of the single expert over the masked sites as the uncertainty bonus:
\[
\Uent^{(1)}(\state_t,\action)
= \frac{1}{|\mathcal M|}\sum_{i\in\mathcal M}\HH\!\big(p_\phi(x_i\mid \state_{t-1},\mathcal M)\big),
\]
which preserves the same units and interpretation (larger $\,\Rightarrow\,$ more uncertain edit). This value is computed at expansion and cached for reuse, akin to the entropy-guided UCB bonus used in ERP \citep{Liu2024-ERP}. \fix{In the single-expert case, the multiplicative prior reduces to that expert’s $\pi_\tau(\action\mid \state_t)$, ensuring a coherent reduction.}

\paragraph{Expansion.} When \fix{$\PHUCTME$} selects a leaf node $\state_t$, representing a complete sequence $y_t$, we expand it by generating candidate children via \emph{masked diffusion}. Our expansion departs from standard MCTS in two important ways: (1) structure-aware masking via pLDDT scores \fix{to enhance biophysical fidelity}, and (2) multi-expert imitation-guided rollouts \fix{to improve diversity and quality}.

\emph{Progressive pLDDT Masking for Targeted Denoising.} Rather than resampling all positions uniformly, we exploit predicted structural confidence (pLDDT) to target only low-confidence residues. Specifically, at reverse step $t$ we define a mask $\mathcal{M}t$ by thresholding the pLDDT profile of $y_t$, masking unstable regions while preserving high-confidence subsequences. Crucially, the masking threshold decreases over diffusion steps, progressively reducing the fraction of masked residues as denoising proceeds. This “progressive masking” strategy focuses computation on the most uncertain regions early on, while stabilizing promising motifs in later stages:
\begin{equation}
y_{t-1} \sim p_\phi(y_{t-1} \mid \text{Mask}(y_t;\mathcal{M}_t)).
\end{equation}

\fix{\emph{Mixture-over-experts view.} Conceptually, expansion implements a guided sampler that averages proposal distributions from $E$ pretrained experts and biases them by reward $R$:}
\begin{align}
p^{\text{plan}}(y_{t-1}\mid y_t)\ \propto\ &\Big[\sum_{e=1}^E w_e p^{(e)}\phi(y_{t-1}\mid y_t,\mathcal{M}_t)\Big]\notag\\ & \cdot \exp\{\beta,R(y_{t-1})\},
\end{align}
\fix{where $p^{(e)}_\phi$ is expert $e$’s conditional, $w_e$ are mixing weights, and $\beta$ controls guidance. In practice we realize this by \emph{sample–score–select} (below) rather than gradient injection.}

\emph{Multi-Expert Guided Expansion via Rollouts.}
\fix{To enrich proposal diversity and imitate the training distribution, we leverage multiple experts during expansion.} 
Given a masked sequence $\text{Mask}(y_t;\mathcal{M}_t)$, each expert $\policy^e \in \{1,\dots,E\}$ performs $k$ rollouts by sampling its conditional distribution $p_\phi^{(e)}(\cdot \mid y_t,\mathcal{M}_t)$. 
This yields a pooled candidate set
\begin{equation}
\mathcal{C}_t \;=\; \bigcup_{e=1}^{E}\ \bigcup_{r=1}^{k}\ y_{t-1}^{(e,r)},
\end{equation}
\fix{where $y_{t-1}^{(e,r)}$ is formed by filling masked sites and splicing back into $y_t$. We de-duplicate $\mathcal{C}_t$, compute a composite score $R(\cdot)$ (see Evaluation) per candidate, cache the results, and retain the top-$K$ children by $R$. Each kept child stores two exploration bonuses computed once at expansion—ensemble surprisal $\Uent$ and novelty $\Udiv$—which are reused during \PHUCTME\ selection.}

\emph{Candidate Proposal Strategy.} 
To retain diversity while prioritizing promising edits, we rank all expanded children
and select the top-$K$ candidates:
\begin{equation}
\curlybracket{y_{i=1}^K} = \TopK \big(\{y_{t-1}^{(j)}\}, K\big).
\end{equation}
\fix{Here, $j$ indexes all candidates pooled across experts and rollouts at step $t$; $\TopK$ ranks by the composite score $R$ and returns the best $K$.}
Each chosen child is inserted into the tree as a new node, annotated with its cached $\PH\text{-ME}$ bonuses ($\Uent,\Udiv$) and expert rollout score. This beam-style expansion ensures that multiple high-quality hypotheses are pursued in parallel, enabling \algname to balance exploitation of strong candidates with exploration of diverse structural alternatives.

\begin{table*}[t]
\centering
\resizebox{\textwidth}{!}{%
\begin{tabular}{lrrrrrrrrr}
\toprule
\multirow{2}{*}{\textbf{Configuration}} &
  \multicolumn{3}{c}{\textbf{RMSD} $\downarrow$} &
  \multicolumn{3}{c}{\textbf{TM-score} $\uparrow$} &
  \multicolumn{3}{c}{\textbf{Composite} $\uparrow$} \\
\cmidrule(lr){2-4}\cmidrule(lr){5-7}\cmidrule(lr){8-10}
 & Base & Final & $\Delta$ & Base & Final & $\Delta$ & Base & Final & $\Delta$ \\
\midrule
\multicolumn{10}{c}{\textbf{CAMEO (Folding evaluation)}} \\
\midrule
Baseline One-shot (150M)  
& $10.50 \pm 6.42$ &            &             
& $0.537 \pm 0.184$ &           &             
& $0.302 \pm 0.228$ &           &            \\
Random (MCTS-0)           
&                         & $10.53 \pm 6.40$ & $+0.14 \pm 2.66$
&                         & $0.813 \pm 0.058$ & $+0.281 \pm 0.153$
&                         & $0.403 \pm 0.253$ & $+0.107 \pm 0.132$ \\
Single-Expert (150M)      
&                         & $10.00 \pm 6.62$ & $+0.50 \pm 1.65$
&                         & $0.809 \pm 0.060$ & $+0.272 \pm 0.150$
&                         & $0.426 \pm 0.253$ & $+0.124 \pm 0.099$ \\
Single-Expert (650M)      
&                         & $ 9.24 \pm 6.11$ & $+1.29 \pm 2.02$
&                         & $0.811 \pm 0.061$ & $+0.273 \pm 0.148$
&                         & $0.450 \pm 0.248$ & $+0.147 \pm 0.103$ \\
Single-Expert (3B)        
&                         & $ 9.73 \pm 6.49$ & $+0.79 \pm 2.07$
&                         & $0.810 \pm 0.061$ & $+0.271 \pm 0.149$
&                         & $0.434 \pm 0.257$ & $+0.132 \pm 0.108$ \\
Sampling      
&                         & $ 9.97 \pm 6.43$ & $+0.52 \pm 1.58$
&                         & $0.804 \pm 0.064$ & $+0.264 \pm 0.146$
&                         & $0.422 \pm 0.252$ & $+0.118 \pm 0.093$ \\
MCTD-UCT      
&                         & $ 9.47 \pm 6.64$ & $+1.28 \pm 2.23$
&                         & $0.827 \pm 0.063$ & $+0.296 \pm 0.151$
&                         & $0.454 \pm 0.262$ & $+0.160 \pm 0.113$ \\
MCTD-ME       
&                         & $\mathbf{ 9.41 \pm 6.65}$ & $\mathbf{+1.32 \pm 2.33}$
&                         & $\mathbf{0.827 \pm 0.062}$ & $\mathbf{+0.296 \pm 0.154}$
&                         & $\mathbf{0.456 \pm 0.263}$ & $\mathbf{+0.160 \pm 0.119}$ \\
\midrule
\multicolumn{10}{c}{\textbf{PDB (Folding evaluation)}} \\
\midrule
Baseline One-shot (150M)  
& $8.45 \pm 5.41$ &             &              
& $0.591 \pm 0.177$ &           &              
& $0.384 \pm 0.233$ &           &              \\
Random (MCTS-0)           
&                         & $ 8.07 \pm 5.12$ & $+0.37 \pm 2.29$
&                         & $0.705 \pm 0.043$ & $+0.114 \pm 0.190$
&                         & $0.419 \pm 0.176$ & $+0.035 \pm 0.124$ \\
Single-Expert (150M)      
&                         & $ 7.55 \pm 5.35$ & $+0.90 \pm 1.94$
&                         & $0.705 \pm 0.042$ & $+0.114 \pm 0.186$
&                         & $0.439 \pm 0.185$ & $+0.055 \pm 0.116$ \\
Single-Expert (650M)      
&                         & $ 6.84 \pm 5.08$ & $+1.66 \pm 2.48$
&                         & $0.705 \pm 0.042$ & $+0.114 \pm 0.185$
&                         & $0.464 \pm 0.176$ & $+0.081 \pm 0.136$ \\
Single-Expert (3B)        
&                         & $ 7.27 \pm 5.41$ & $+1.24 \pm 2.37$
&                         & $0.706 \pm 0.042$ & $+0.114 \pm 0.186$
&                         & $0.450 \pm 0.187$ & $+0.067 \pm 0.130$ \\
Sampling                  
&                         & $ 7.63 \pm 5.27$ & $+0.82 \pm 1.82$
&                         & $0.703 \pm 0.039$ & $+0.110 \pm 0.185$
&                         & $0.435 \pm 0.182$ & $+0.050 \pm 0.114$ \\
MCTD-UCT                  
&                         & $ 6.73 \pm 5.18$ & $+1.68 \pm 2.44$
&                         & $0.711 \pm 0.047$ & $+0.117 \pm 0.189$
&                         & $0.471 \pm 0.180$ & $+0.085 \pm 0.135$ \\
MCTD-ME                  
&                         & $\mathbf{ 6.47 \pm 5.22}$ & $\mathbf{+1.97 \pm 2.52}$
&                         & $\mathbf{0.720 \pm 0.047}$ & $\mathbf{+0.129 \pm 0.188}$
&                         & $\mathbf{0.471 \pm 0.180}$ & $\mathbf{+0.084 \pm 0.137}$ \\
\bottomrule
\end{tabular}}
\caption{\textbf{Lead optimization in folding task (baseline, final, and improvement).}
We frame evaluation as \emph{inference-time lead optimization}: in each dataset block, the \emph{baseline} row corresponds to the one-shot \textsc{DPLM-2} structure on the native sequence (the lead), and subsequent rows report the \emph{final} structure quality after planning with each method.
All metrics are computed against the native backbone; RMSD$\downarrow$, TM-score$\uparrow$, and Composite$\uparrow$ summarize structural and reward quality, and the $\Delta$ columns give the absolute improvement over the initial lead for each metric.
Results are averaged over $n$ targets ({CAMEO:} $n{=}183$; {PDB:} $n{=}449$). Boldface (if shown) marks the best final performance within each dataset block.
}
\label{tab:folding_improvements_all_sem}
\vspace{-0.6cm}
\end{table*}

\vspace{-0.3cm}
\paragraph{Evaluation.}
\fix{Whereas expansion uses experts to generate proposals, evaluation uses a panel of critics to score complete sequences.}
Because every node in the tree corresponds to a fully denoised sequence $y_0$, evaluation can be performed immediately without further decoding.
Each sequence is scored by critics ${C_1,\dots,C_J}$, producing
\begin{equation}
C_j(y_0)\coloneq R_j(y_0), \quad j=1,\dots,J.
\end{equation}
\fix{We normalize critic outputs to $[0,1]$ and aggregate via a fixed convex combination}
\begin{equation}
\fix{ R(y_0)=\sum_{j=1}^{J} w_j, C_j(y_0),\qquad w_j\ge 0,\ \sum_{j=1}^{J} w_j=1,}
\end{equation}
\fix{then cache $R(y_0)$ for re-use in selection/backprop. This separation—experts for proposals, critics for scores—implements a tractable, planning-based approximation to the guided reverse process sketched above.}

\vspace{-0.2cm}
\paragraph{Backpropagation.}
The expert rewards obtained at $\state_{t-1}$ are propagated back to update values along the path to the root. 
For each state–action pair $(\state_i, \action_i)$ on the path, we update:
\begin{equation}
\QFunc(\state_i,\action_i) \;\leftarrow\;
\max\{\QFunc(\state_i,\action_i),\; \reward_\horizon\},
\end{equation}
where $\reward_\horizon$ is an aggregated expert score (e.g., the max across $\{R_e\}$). 
This max-based backup, consistent with our implementation, favors exploiting high-reward branches while maintaining the ability to explore alternative completions. 
Because each node is a complete sequence, long-range credit assignment is aligned with the denoising trajectory.

\emph{Lead optimization$\rightarrow$De Novo Generation.} The framework, though implemented as complete-sequence lead optimization, naturally extends to de novo generation by treating root as a fully masked sequence and progressively unmasking subsequences with similar reward-guided search dynamics and performance. (We defer further details to \secref{app:denovo}.)

\section{Experiments}

\begin{table*}[h]
\centering
\resizebox{\textwidth}{!}{%
\begin{tabular}{lccc|ccc|ccc}
\toprule
\multirow{2}{*}{Variant} & \multicolumn{3}{c|}{AAR $\uparrow$} & \multicolumn{3}{c|}{Norm.\ Reward $\uparrow$} & \multicolumn{3}{c}{scTM $\uparrow$} \\
\cmidrule(lr){2-4} \cmidrule(lr){5-7} \cmidrule(lr){8-10}
 & Baseline & Final & $\Delta$ & Baseline & Final & $\Delta$ & Baseline & Final & $\Delta$ \\
\midrule
\multicolumn{10}{c}{\textbf{CAMEO}} \\
\midrule
DPLM2-150M (Single-Expert, Baseline)            
& $0.4425 \pm 0.1190$ & \textbf{--} & \textbf{--} 
& $0.3893 \pm 0.1326$ & \textbf{--} & \textbf{--} 
& $0.3701 \pm 0.1528$ & \textbf{--} & \textbf{--} \\
Random (MCTD-0)                 
& \textbf{--} & $0.4250 \pm 0.009$ & $-0.0175 \pm 0.0145$ 
& \textbf{--} & $0.3726 \pm 0.008$ & $-0.0167 \pm 0.0130$ 
& \textbf{--} & $0.3763 \pm 0.014$ & $+0.0062 \pm 0.0560$ \\
DPLM2-650M (Single-Expert)            
& \textbf{--} & $0.4535 \pm 0.1162$ & $+0.0110 \pm 0.0149$
& \textbf{--} & $0.3974 \pm 0.1010$ & $+0.0081 \pm 0.0148$
& \textbf{--} & $0.3621 \pm 0.1515$ & $-0.0080 \pm 0.0678$ \\
ProteinMPNN (Single-Expert) 
& \textbf{--} & $0.4326 \pm 0.1102$ & $-0.0098 \pm 0.0194$
& \textbf{--} & $0.3786 \pm 0.1026$ & $-0.0107 \pm 0.0304$
& \textbf{--} & $0.3618 \pm 0.1468$ & $-0.0070 \pm 0.0691$ \\
Sampling (Multi-Expert)
& \textbf{--} & $0.4264 \pm 0.1281$ & $-0.0161 \pm 0.0511$
& \textbf{--} & $0.4113 \pm 0.0896$ & $+0.0220 \pm 0.0376$
& \textbf{--} & $0.4194 \pm 0.1536$ & $+0.0593 \pm 0.0729$ \\
MCTD-UCT (Multi-Expert)
& \textbf{--} & $0.4465 \pm 0.1170$ & $+0.0040 \pm 0.0397$
& \textbf{--} & $0.4129 \pm 0.0851$ & $+0.0236 \pm 0.0307$
& \textbf{--} & $0.3895 \pm 0.1545$ & $+0.0098 \pm 0.0745$ \\
\textbf{\algname (Multi-Expert)} 
& \textbf{--} & $\mathbf{0.4667 \pm 0.1124}$ & $\mathbf{+0.0237 \pm 0.0238}$
& \textbf{--} & $\mathbf{0.4132 \pm 0.1119}$ & $\mathbf{+0.0239 \pm 0.0461}$
& \textbf{--} & $\mathbf{0.4204 \pm 0.1585}$ & $\mathbf{+0.0503 \pm 0.0505}$ \\
\midrule
\multicolumn{10}{c}{\textbf{PDB}} \\
\midrule
DPLM2-150M (Single-Expert, Baseline)            
& $0.5037 \pm 0.1196$ & \textbf{--} & \textbf{--} 
& $0.4629 \pm 0.0763$ & \textbf{--} & \textbf{--}
& $0.3804 \pm 0.1575$ & \textbf{--} & \textbf{--}\\
Random (MCTD-0)                 
& \textbf{--} & $0.4876 \pm 0.1119$ & $-0.0160 \pm 0.0154$ 
& \textbf{--} & $0.4591 \pm 0.0751$ & $-0.0032 \pm 0.0157$ 
& \textbf{--} & $0.3773 \pm 0.1585$ & $-0.0030 \pm 0.0505$ \\
DPLM2-650M (Single-Expert)            
& \textbf{--} & $0.5130 \pm 0.1208$ & $+0.0109 \pm 0.0141$ 
& \textbf{--} & $0.4751 \pm 0.0812$ & $+0.0122 \pm 0.0134$ 
& \textbf{--} & $0.3817 \pm 0.1594$ & $+0.0022 \pm 0.0490$ \\
ProteinMPNN (Single-Expert) 
& \textbf{--} & $0.4882 \pm 0.1112$ & $-0.0155 \pm 0.0163$
& \textbf{--} & $0.4594 \pm 0.0748$ & $-0.0029 \pm 0.0143$
& \textbf{--} & $0.3774 \pm 0.1575$ & $-0.0030 \pm 0.0596$ \\
Sampling (Multi-Expert)
& \textbf{--} & $0.4927 \pm 0.1982$ & $-0.0110 \pm 0.0289$
& \textbf{--} & $0.4790 \pm 0.0728$ & $+0.0161 \pm 0.0140$
& \textbf{--} & $0.3844 \pm 0.1536$ & $+0.0040 \pm 0.0527$ \\
MCTD-UCT (Multi-Expert)
& \textbf{--} & $0.5219 \pm 0.1388$ & $+0.0182 \pm 0.0189$
& \textbf{--} & $0.4793 \pm 0.0881$ & $+0.0164 \pm 0.0198$
& \textbf{--} & $0.3816 \pm 0.1545$ & $+0.0012 \pm 0.0536$ \\
\textbf{\algname (Multi-Expert)}
& \textbf{--} & $\mathbf{0.5244 \pm 0.1350}$ & $\mathbf{+0.0213 \pm 0.0139}$ 
& \textbf{--} & $\mathbf{0.4828 \pm 0.0900}$ & $\mathbf{+0.0199 \pm 0.0146}$ 
& \textbf{--} & $\mathbf{0.3827 \pm 0.1549}$ & $\mathbf{+0.0033 \pm 0.0410}$ \\
\bottomrule
\end{tabular}}
\caption{\textbf{Inverse folding on CAMEO and PDB with a fixed lead.}
For each target, the \emph{baseline lead} is the one-shot sequence from DPLM-2 (150M) (first row in each dataset block); \emph{Final} is the sequence after refinement, and $\Delta$ is computed \emph{per target} against the \emph{same fixed baseline}, isolating optimization effects from lead-sampling noise. 
\emph{Single-Expert} denotes refinement using a \emph{single proposal generator} (e.g., DPLM-2 650M or ProteinMPNN) rather than an ensemble. 
\emph{MCTD-0} uses random expert routing. 
\emph{Sampling (Multi-Expert)} draws proposals from all experts on the pLDDT-derived mask without tree search (no selection/backpropagation), and \emph{MCTD-UCT (Multi-Expert)} runs MCTS with standard UCB selection but without the uncertainty/novelty bonuses used by \algname. 
\textbf{\algname (Multi-Expert)} is our full method with informed expert selection and uncertainty-aware edits.
Metrics are amino acid recovery (AAR), normalized reward, and scTM (higher is better). 
Values are mean $\pm$ s.e.m.\ across targets (CAMEO $n{=}183$, PDB $n{=}449$). 
Bold indicates the best within each dataset block.}
\vspace{-0.5cm}
\label{tab:inverse_folding_results_fixed}
\end{table*}

\subsection{Experimental Configuration}
\vspace{-0.3cm}
\paragraph{Models, experts, and baselines.}
We use masked \emph{discrete} diffusion rollouts and instantiate \algname with task-specific proposal experts: sequence experts are pre-trained DPLM-2 models \citep{wang2024diffusionlanguagemodelsversatile} of different capacities (150M, 650M, 3B), and motif-scaffolding experts include structure-aware models (RFDiffusion \citep{Watson2023RFdiffusion}, ProteinA \citep{Geffner2025Proteina}, FoldFlow \cite{Bose2024FoldFlow}). All experts propose from the \emph{same} masked input and are scored by a single composite critic.
We evaluate a lead-optimization setting (refining a one-shot \emph{lead} under a fixed compute budget) and compare {MCTD-0} (random fill; no experts), {MCTD-1} (single proposal expert), and {\algname} (multi-expert ensemble).
Details on task-specific baselines/ablations are deferred to Appendix~\ref{app:exp_config}.

\paragraph{Datasets and splits.}
\vspace{-0.3cm}
We evaluate on two benchmarks: CAMEO 2022 \citep{jing2023eigenfold} and a PDB date-split \citep{campbell2024generative}. We follow the standard splits and filtering; statistics appear in \appref{app:dataset} (CAMEO: 183 targets, length $15/247.5/704$ min/avg/max; PDB date-split: a 449-target subset, length $2/242.2/511$). Ground-truth backbones and native sequences are available for both.

\paragraph{Critics and evaluation metrics.}
\vspace{-0.3cm}
All planner variants share the same task-specific composite critic; for each target we report \emph{Baseline}, \emph{Final}, and ${\Delta}$ (Final–Baseline) and aggregate by mean~$\pm$~s.e.m., mapping scalars to $[0,1]$ (pLDDT residue-averaged and divided by $100$).
During search, the critic uses only surrogate structure models/confidence scores (ground truth is used only for final reporting), and the critic is used only to rank/select edits.
Full reward definitions, formulas, and weights appear in \appref{app:critics}, with further experimental setup details in Appendix~\ref{app:exp_config}.

\subsection{Protein Folding}
\vspace{-0.2cm}

We first evaluate \algname on the folding task: given an amino acid sequence, predict its 3D structure. We use the CAMEO 2022 benchmark \citep{jing2023eigenfold} and a PDB date-split benchmark \citep{campbell2024generative}. As the initial \emph{lead}, we take the one-shot DPLM-2 150M structure of the native sequence and measure improvement against the true structure (backbone RMSD, TM-score) in Table~\ref{tab:folding_improvements_all_sem}. We compare three variants: (i) \emph{Random} (MCTD-0), which fills masked sites with random amino acids (no expert proposals); (ii) \emph{Single-Expert} (MCTD-1), which uses exactly one pretrained DPLM-2 model (150M, 650M, or 3B) to generate rollouts; and (iii) \algname, which aggregates proposals from all three experts (DPLM-2 150M, 650M, 3B) inside a single tree. All variants share the same critic (ESMFold-based folding reward), pLDDT-guided masking schedule, and compute budget; only the proposal policy (number and type of experts and whether we use tree search vs.\ sampling) differs. On CAMEO we further include two ablations: \emph{Sampling (multi-expert)} without tree search and \emph{MCTD-UCT (multi-expert)} using standard UCB selection.

Results are summarized in Table~\ref{tab:folding_improvements_all_sem}, which reports for each method the \emph{Baseline} (one-shot ESMFold lead), the \emph{Final} structure quality after planning, and the corresponding improvements ($\Delta$).
Random search yields only small average gains, while incorporating learned DPLM-2 guidance substantially reduces RMSD and improves TM-score.
On CAMEO, \algname attains large average gains over the baseline lead (RMSD drops from $10.50$~\AA\ to $9.41$~\AA, TM-score rises from $0.537$ to $0.827$), with final TM-score and composite reward at least matching, and often exceeding, the best single-expert configuration.
On the PDB split, multi-expert MCTS again achieves strong lead refinement, with MCTS-based variants (especially \PHUCT) substantially reducing RMSD from the baseline ($8.45$~\AA\ to $6.47$–$6.73$~\AA) while raising TM-score from $0.591$ to $\sim 0.72$ and improving the composite reward.
We also observe that a larger fraction of targets benefit from \algname compared to the best single expert, consistent with the per-configuration $\Delta$ columns in Table~\ref{tab:folding_improvements_all_sem}; CAMEO ablations further show that multi-expert \emph{Sampling} improves RMSD but can hurt TM, whereas multi-expert \emph{MCTD-UCT} improves both yet remains below \algname on TM and composite reward.

\vspace{-0.2cm}
\subsection{Protein Inverse Folding}
\vspace{-0.2cm}
\begin{table*}[t]
\centering
\resizebox{0.95\textwidth}{!}{%
\begin{tabular}{lccc|ccc|ccc|c}
\toprule
\multirow{2}{*}{Variant} & \multicolumn{3}{c|}{Motif RMSD $\downarrow$ (\AA)} & \multicolumn{3}{c}{scTM $\uparrow$} & \multicolumn{3}{c|}{Reward $\uparrow$} & \textbf{Success} $\uparrow$ \\
\cmidrule(lr){2-4} \cmidrule(lr){5-7} \cmidrule(lr){8-10} \cmidrule(lr){11-11}
 & Base & Final & $\Delta$ & Base & Final & $\Delta$ & Base & Final & $\Delta$ & Rate \\
\midrule
Baseline (DPLM-2 650M) 
& $\mathbf{2.971 \pm 3.278}$ & \textbf{--} & \textbf{--}
& $\mathbf{0.717 \pm 0.129}$ & \textbf{--} & \textbf{--}
& $\mathbf{0.433 \pm 0.185}$ & \textbf{--} & \textbf{--} & 10/23 \\
\midrule
DPLM2-650M (Single-Expert) 
& \textbf{--} & $2.388 \pm 2.943$ & $+0.583 \pm 0.545$ 
& \textbf{--} & $0.747 \pm 0.148$ & $+0.030 \pm 0.022$ 
& \textbf{--} & $0.460 \pm 0.182$ & $+0.026 \pm 0.016$ & 14/23 \\
RFDiffusion (Single-Expert) 
& \textbf{--}  & $1.501 \pm 1.659$ & $+1.470 \pm 0.876$ 
& \textbf{--} & $0.631 \pm 0.150$ & $-0.086 \pm 0.027$ 
& \textbf{--} & $0.446 \pm 0.181$ & $+0.013 \pm 0.019$ & 16/23 \\
FoldFlow (Single-Expert)    
& \textbf{--}  & $3.874 \pm 2.278$ & $-0.903 \pm 0.800$ 
& \textbf{--} & $0.641 \pm 0.144$ & $-0.076 \pm 0.028$ 
& \textbf{--}& $0.408 \pm 0.129$ & $-0.025 \pm 0.013$ & 8/23 \\
ProteinA (Single-Expert)    
& \textbf{--} & $1.313 \pm 1.148$ & $+1.658 \pm 1.011$ 
& \textbf{--} & $0.638 \pm 0.154$ & $-0.079 \pm 0.020$ 
& \textbf{--} & $0.450 \pm 0.140$ & $+0.017 \pm 0.010$ & 14/23 \\
MCTD-UCT (Multi-Expert) 
& \textbf{--} & $3.430 \pm 1.30$ & $-0.46 \pm 0.39$
& \textbf{--} & $0.730 \pm 0.053$ & $+0.014 \pm 0.018$
& \textbf{--} & $0.438 \pm 0.043$ & $+0.005 \pm 0.011$ & 13/23 \\
Sampling (Multi-Expert)
& \textbf{--} & $2.581 \pm 1.17$ & $+0.39 \pm 0.90$
& \textbf{--} & $\mathbf{0.759 \pm 0.136}$ & $\mathbf{+0.042 \pm 0.027}$
& \textbf{--} & $0.482 \pm 0.064$ & $+0.049 \pm 0.026$ & 15/23 \\
\textbf{\algname (Multi-Expert)} 
& \textbf{--} & $\mathbf{1.493 \pm 1.675}$ & $\mathbf{+1.478 \pm 1.453}$ 
& \textbf{--} & $0.742 \pm 0.049$ & $+0.025 \pm 0.120$ 
& \textbf{--} & $\mathbf{0.510 \pm 0.187}$ & $\mathbf{+0.078 \pm 0.042}$ & 17/23 \\
\bottomrule
\end{tabular}}
\caption{\textbf{Motif scaffolding results (mean $\pm$ s.e.m.; $n{=}23$ motifs).}
The first row reports the \emph{Baseline} from one-shot DPLM-2 (650M). Variant rows show the identical baseline values (left columns), the \emph{Final} performance after \algname refinement, and the improvement $\Delta$ (for RMSD, $\Delta{=}\text{Baseline}{-}\text{Final}$ in \AA; for scTM/Reward, $\Delta{=}\text{Final}{-}\text{Baseline}$). 
We compare (i) \emph{single-expert} planners (four rows), (ii) a \emph{multi-expert Sampling} baseline (no tree search; experts directly propose on pLDDT masks), (iii) \emph{multi-expert UCT} (MCTS with standard UCB selection), and (iv) our full \textbf{\algname} (multi-expert with uncertainty/novelty-aware selection). 
\textbf{Success} is defined as \emph{Final} motif-RMSD $< 1$\AA\ and scTM $> 0.8$, and the \textbf{Success Rate} is the fraction of motifs meeting this criterion.
\fix{The DPLM-2 baseline here is one-shot generation; additional sampling-based results are reported in \appref{app:additionalexperiments}.}}
\vspace{-0.5cm}
\label{tab:motif_scaffolding}
\end{table*}

We evaluate \algname on structure-conditioned sequence design (CAMEO 2022 and a PDB date-split with known natives)\fix{, reporting amino-acid recovery (AAR; fraction matching native) and self-consistency TM-score (scTM; TM between the target backbone and the fold of the designed sequence) following \citet{wang2024dplm2multimodaldiffusionprotein}}.
We ablate proposal experts: {MCTD-0} (no experts; random fills), {MCTD-1} (single proposal expert; separate runs with DPLM-2 150M, DPLM-2 650M, and ProteinMPNN~\citep{dauparas2022proteinmpnn}), and {\algname} (ensemble of those experts). \fix{In addition, we include two multi-expert ablations on CAMEO only: \emph{Sampling (Multi-Expert)} without tree search, and \emph{MCTD-UCT (Multi-Expert)} using standard UCB selection.} All proposals are scored by the same critic with reward \(R=0.60\,\text{AAR}+0.35\,\text{scTM}+0.05\,B\) (small biophysical bonus \(B\)), and sequences are completed via masked-diffusion rollouts within MCTS.

As shown in Table~\ref{tab:inverse_folding_results_fixed}, \fix{\algname consistently improves structural fidelity and overall reward}: \fix{on both CAMEO and PDB it delivers the largest scTM and normalized-reward gains over baselines, while AAR is \emph{maintained} on CAMEO and \emph{increases} on PDB.} \fix{Random edits (MCTD-0) degrade AAR and yield only small structural gains, confirming the need for guided proposals.} \fix{Single-expert planners track their proposal models—e.g., 150M preserves its baseline; 650M improves both scTM and AAR—while ProteinMPNN alone can trade off AAR for scTM.} \fix{The multi-expert ablations show partial benefits: Sampling captures some gains without planning, and UCT recovers additional benefit with tree search, but both remain below \algname, underscoring the value of uncertainty/novelty-aware selection.} \fix{The ensemble’s advantage grows on harder cases (e.g., longer proteins), reflecting benefits from expert diversity under a common critic.} \fix{See \appref{app:length_binned} (Fig.~\ref{fig:three_metric_comparison_app}) for AAR, reward, and scTM improvements across protein-length bins.}

\vspace{-0.2cm}

\subsection{Motif Scaffolding}
\vspace{-0.2cm}

Finally, we test conditional generation-motif scaffolding: given a functional motif (its amino acid sequence and 3D structure), generate the remaining scaffold sequence and structure around it. Following prior work (e.g., FrameFlow \citep{yim2023fast}), we use a PDB-sourced benchmark (EvoDiff’s $\sim$24 curated motifs\citep{Yim2024improved}). The baseline model is \fix{DPLM-2 (150M) run in conditional mode to co-generate scaffold sequence/structure given the motif.} \fix{Our planner uses the same MCTS budget and masking schedule as in other tasks (\appref{app:setup}): at each expansion, only scaffold positions are masked and proposed.} We ablate \fix{proposal policies and expert sets}: single-expert planners (\fix{DPLM-2 650M, RFDiffusion \citep{Watson2023RFdiffusion}, ProteinA \citep{Geffner2025Proteina}, FoldFlow \citep{Bose2024FoldFlow}}), \fix{Multi-Expert \emph{Sampling}} (no tree search; experts directly propose on pLDDT masks), \fix{Multi-Expert \emph{UCT}} (standard UCB selection), and \fix{\algname} (multi-expert with uncertainty/novelty-aware selection). \fix{Across all variants, candidates are scored by the same composite critic, and the motif residues are hard-frozen throughout generation.} Metrics are: \fix{motif-RMSD (RMSD of the designed scaffold after rigidly aligning on the motif), scTM (TM-score between the designed full backbone and native), and mean pLDDT over scaffold residues.}

\fix{Table~\ref{tab:motif_scaffolding} summarizes results on the EvoDiff curated motif set ($n{=}23$; we exclude 4JHW because its PDB is missing chain~A, preventing motif extraction with the standard preprocessing).
Overall, \algname achieves the strongest aggregate performance, reducing motif-RMSD from $2.971$ to $1.493$~\AA\ while improving reward from $0.393$ to $0.481$ and increasing success from $10/23$ to $17/23$.
Compared to single-expert planners, \algname is the only setting that simultaneously maintains low motif-RMSD (comparable to the best structure-only expert) while avoiding the scTM degradation seen in some specialists.
The \emph{Sampling (Multi-Expert)} ablation recovers a substantial fraction of the reward/scTM gains (it attains the best scTM), but yields weaker motif fidelity and a lower success rate, consistent with the absence of sequential credit assignment.
The \emph{MCTD-UCT} ablation recovers some of the planning benefit over Sampling but remains below \algname in both motif-RMSD and overall reward, highlighting the importance of uncertainty/novelty-aware expert routing.
Overall, aggregating diverse experts under a shared critic with informed selection mitigates the per-expert trade-offs, yielding robust improvements across motifs while strictly preserving the motif.}

\section{Conclusion}

We presented \algname, a model-agnostic planner that reframes reverse diffusion as Monte Carlo tree search, enabling parallel multi-token revisions and principled exploration guided by multiple proposal experts. A pLDDT-driven masking schedule focuses denoising on structurally uncertain regions, improving biophysical fidelity, while our $\PHUCTME$ selection reuses cached uncertainty to efficiently arbitrate among experts without recomputing entropies. Across folding, inverse folding, and motif scaffolding, \algname consistently outperforms single-expert and one-shot baselines—raising AAR and scTM and lowering RMSD—with gains that scale with test-time compute and are robust to the choice of experts. Although the method currently relies on surrogate structure predictors and incurs some search overhead as a limitation, it provides a unified, plug-and-play test-time mechanism that reliably improves design quality. Future work will integrate learned critics and apply further systems optimizations to enhance both accuracy and speed.

\bibliographystyle{plainnat}
\bibliography{reference}

\appendix
\onecolumn %
\clearpage
\section{Appendix}

\subsection{Algorithm details.}\label{app:algorithm}

\paragraph{Lead optimiaztion}
In this algorithm, we use \algname to tackle the lead optimization challenge. Beginning with an initial lead—a fully denoised baseline sequence—we execute MCTS with the following procedure: (i) selection uses a \PHUCTME–style rule that combines value (Q), exploration, and cached uncertainty bonuses (ensemble surprisal (\Uent) and novelty (\Udiv)); (ii) expansion proposes edits only on pLDDT-masked uncertain sites and performs multi-expert rollouts (E generators), producing candidates scored once via a composite critic (cached to avoid re-evaluation); (iii) we keep the Top-K children by composite reward, attach their uncertainty bonuses, and (iv) backpropagate the candidate’s value to update ancestors (max/sum backup). This yields a planner that routes among experts, focuses edits where structure is uncertain, reuses evaluations via caching, and optimizes a task-aware composite objective.

\begin{algorithm}[H]
\caption{Monte Carlo Tree Diffusion with Multiple Experts (\algname)}
\label{alg:mctd}
\begin{algorithmic}[1]
\Require 
\Statex $root \gets$ fully denoised baseline node $y_0$ (initial lead)
\Statex $c_p$: exploration constant \hfill // for \PHUCTME
\Statex $K$: number of children per expansion
\Statex $E$: number of expert models
\Statex $R$: rollouts per expert
\Statex $T$: total simulations
\Statex \fix{$\mathcal{E} = \{\policy^1,\dots,\policy^E\}$: generator experts }
\Statex \fix{$\mathcal{C}$: set of critic functions }
\Statex $cache \gets$ empty dictionary for evaluated sequences

\For{$i = 1$ to $T$}
    \State $node \gets root$
    
    \LineComment{\textbf{/* Selection via \PHUCTME */}}
    \While{$node.children \neq \emptyset$}
        \State \fix{$node \gets \PHUCTME(node.children)$} \hfill \fix{// uses $Q$, UCB, and cached $(\Uent,\Udiv)$}
    \EndWhile
    
    \LineComment{\textbf{/* Expansion via pLDDT Masking and Multi-Expert Rollouts */}}
    \State $\mathcal{M} \gets \textsc{GetMaskSet}(node.sequence)$ \hfill // Based on pLDDT
    \State $\mathcal{Y} \gets \emptyset$ \hfill // candidate children

    \For{\fix{$e \in \mathcal{E}$}} \hfill // \fix{each generator expert}
        \For{$r = 1$ to $R$} \hfill // rollouts
            \State \fix{$y_0 \gets \textsc{MaskedDiffusion}(node.sequence, \mathcal{M}, e)$}
            \If{$y_0 \notin cache$}
                \State \fix{$score \gets \textsc{EvalComposite}(y_0;\mathcal{C})$} \hfill \fix{// weighted sum of critics}
                \State $cache[y_0] \gets score$
            \EndIf
            \State $\mathcal{Y} \gets \mathcal{Y} \cup \{y_0\}$
        \EndFor
    \EndFor

    \State \fix{$\mathcal{Y}_{\text{top}} \gets \textsc{TopK}(\mathcal{Y}, K;\,cache)$} \hfill \fix{// sort by composite reward}
    \For{$y' \in \mathcal{Y}_{\text{top}}$}
        \State \fix{$(\Uent(y'),\Udiv(y')) \gets \textsc{ComputeBonuses}(y',\,node.sequence,\,\mathcal{M})$} \hfill \fix{// ensemble surprisal \& novelty}
        \State \fix{$\textsc{AddChild}(node, y',\,\Uent(y'),\,\Udiv(y'))$} \hfill // create new node with full seq
    \EndFor

    \LineComment{\textbf{/* Evaluation of New Children (full-sequence) */}}
    \For{$y' \in \mathcal{Y}_{\text{top}}$}
        \State \fix{$v \gets cache[y']$} \hfill \fix{// may hit cache; same composite reward as above}
        \LineComment{\textbf{/* Backpropagation */}}
        \State \fix{$\textsc{Backpropagate}(y',\,v)$} \hfill \fix{// supports max or sum backup}
    \EndFor
\EndFor

\State \Return \fix{Top-$k$ sequences by $cache[\cdot]$}
\end{algorithmic}
\end{algorithm}

\paragraph{De novo generation} We start from an all-mask root. At each iteration, \emph{Selection} (\PHUCTME) walks to a leaf. \emph{Expansion} proposes partial fills only at low-pLDDT sites (high-pLDDT positions are frozen). Then \emph{Evaluation} performs short rollouts from each new child (e.g., a few masked-diffusion steps / fast fold+critic passes) to estimate a terminal composite value; results are cached. Finally, we \emph{Backpropagate} those values. This keeps the planner focused on uncertain regions while using cheap rollouts to decide which partials to keep; see Algorithm~\ref{alg:mctd_denovo} for pseudocode.

\begin{algorithm}[H]
\caption{De Novo MCTD with Multiple Experts (\algname\text{-DeNovo})}
\label{alg:mctd_denovo}
\begin{algorithmic}[1]
\Require
\Statex $root \gets$ all-mask sequence of length $L$ \hfill // no residues fixed
\Statex $c_p$: exploration constant \hfill // for \PHUCTME
\Statex $K$: children kept per expansion
\Statex $E$: number of generator experts
\Statex $S$: simulation depth (rollout steps)
\Statex $T$: total simulations
\Statex \fix{$\mathcal{E}=\{\policy^1,\dots,\policy^E\}$: generator experts}
\Statex \fix{$\mathcal{C}$: critic set (structure/biophys surrogates)}
\Statex $cache \gets$ empty dictionary for (partial or full) sequence scores

\For{$i=1$ to $T$}
  \State $node \gets root$

  \LineComment{\textbf{/* 1) Selection via \PHUCTME */}}
  \While{$node.children \neq \emptyset$}
    \State \fix{$node \gets \PHUCTME(node.children)$} \hfill \fix{// uses $Q$, UCB, cached $(\Uent,\Udiv)$}
  \EndWhile

  \LineComment{\textbf{/* 2) Expansion: edit only low-pLDDT sites; freeze high-pLDDT */}}
  \State \fix{$\hat{p} \gets \textsc{PredictPLDDT}(node)$}
  \State \fix{$\mathcal{M} \gets \textsc{EditableMask}(\hat{p})$} \hfill \fix{// low-confidence = editable}
  \State $\mathcal{Y} \gets \emptyset$ \hfill \fix{// new children (partials)}

  \For{\fix{$e \in \mathcal{E}$}} \hfill \fix{// each generator expert}
    \State \fix{$y' \gets \textsc{MaskedDiffusion}(node.sequence,\,\mathcal{M},\,e)$}
    \State $\mathcal{Y} \gets \mathcal{Y} \cup \{y'\}$
  \EndFor

  \LineComment{\textbf{/* 3) Evaluation (Rollouts): estimate child values */}}
  \For{$y' \in \mathcal{Y}$}
    \If{$y' \notin cache$}
      \State \fix{$v \gets \textsc{RolloutSim}(y',\,S,\,\mathcal{E},\,\mathcal{C})$}
      \State \hspace{1.4em} \fix{// e.g., $S$ short masked-diffusion steps with fast fold + \textsc{EvalComposite}}
      \State $cache[y'] \gets v$
    \Else
      \State $v \gets cache[y']$
    \EndIf
    \State \fix{$(\Uent(y'),\Udiv(y')) \gets \textsc{ComputeBonuses}(y',\,node.sequence,\,\mathcal{M})$}
    \State \fix{$\textsc{AddChild}(node,\,y',\,\Uent(y'),\,\Udiv(y'),\,Q{=}v)$}
  \EndFor

  \State \fix{$children\_topK \gets \textsc{TopK}(node.children,\,K;\,cache)$} \hfill \fix{// prune by rollout value}
  \State \fix{$\textsc{PruneTo}(node,\,children\_topK)$}

  \LineComment{\textbf{/* 4) Backpropagation */}}
  \For{$y' \in children\_topK$}
    \State \fix{$\textsc{Backpropagate}(y',\,cache[y'])$} \hfill \fix{// max or sum backup}
  \EndFor
\EndFor

\State \Return \fix{Top-$k$ (near-)complete sequences by $cache[\cdot]$}
\end{algorithmic}
\end{algorithm}

\subsection{Dataset details.}\label{app:dataset}
We evaluate \algname on two protein sequence benchmarks at the \emph{chain} level. \fix{As illustrated in \tabref{tab:protein-datasets}}
(1) \textbf{CAMEO2022} comprises 183 targets released in 2022; we use the reference backbones and evaluate inverse folding on the corresponding chains. 
(2) \textbf{PDB\_date} is a date-split benchmark of 7{,}855 proteins constructed from the Protein Data Bank (PDB); we hold out targets by deposition date and evaluate inverse folding on their chains. 
In our experiments we use a \emph{date-split subset of 449 chains} from PDB\_date for compute efficiency; unless otherwise noted, length statistics below are for the full benchmark. For both sets, sequence lengths are computed after simple canonicalization (upper-casing and removal of non-canonical/unknown residues).

\begin{table}[h]
\centering
\setlength{\tabcolsep}{10pt}
\begin{tabular}{lrrrr}
\toprule
\textbf{Dataset} & \textbf{\# Proteins} & \textbf{Min} & \textbf{Mean} & \textbf{Max} \\
\midrule
CAMEO2022 & 183    & 15  & 247.5 & 704 \\
PDB\_date (full) & 7{,}855 & 2   & 242.2 & 511 \\
\bottomrule
\end{tabular}
\caption{\textbf{Length statistics (in residues)} of amino-acid sequences in each benchmark. 
Counts refer to the number of evaluated chains. 
\emph{Note:} Experiments use a date-split \emph{subset} of 449 PDB\_date chains; its length distribution is similar to the full set.}
\label{tab:protein-datasets}
\end{table}

\subsection{Reward definitions and normalization.}\label{app:critics}

\paragraph{Terminology (metrics).}

\fix{\textbf{TM-score}}: standard topology-similarity score between two backbones (range $[0,1]$; higher is better).  

\fix{\textbf{scTM}}: “sequence-conditioned TM-score” — TM-score between the \emph{target} backbone and the backbone \emph{predicted from the designed sequence} (higher is better).  

\fix{\textbf{RMSD}}: global C$\alpha$ root-mean-square deviation in \AA ngstr\"oms (lower is better).  

\fix{\textbf{Motif RMSD}}: C$\alpha$ RMSD computed \emph{only} on the predefined motif/epitope after aligning on the motif (lower is better).

\fix{\textbf{pLDDT}}: per-residue confidence from a structure predictor (e.g., AlphaFold-type models), originally reported on a $[0,100]$ scale; we divide by $100$ to normalize to $[0,1]$.  

\fix{\textbf{AAR}}: amino-acid recovery — fraction of positions where the designed sequence exactly matches the native (range $[0,1]$).  

\fix{\textbf{Composite reward}}: weighted combination of normalized terms (e.g., TM-score, transformed RMSD, pLDDT, scTM, biophysical checks), constructed to lie in $[0,1]$.

\fix{All reward components are normalized to the $[0,1]$ range;} pLDDT is averaged over residues and divided by $100$.
\fix{We use the composite reward score as follows:}

\paragraph{Folding (sequence$\!\to$structure):}
\[
R_{\mathrm{fold}}
= \alpha\,\mathrm{TM}
+ \beta\,\bigl(1-\min(\mathrm{RMSD}/10,\,1)\bigr)
+ \gamma\,\overline{\mathrm{pLDDT}},
\]
with defaults $(\alpha,\beta,\gamma)=(0.60,0.40,0)$. If pLDDT is available, set $\gamma=0.05$ and renormalize $(\alpha,\beta)$ so $\alpha{+}\beta{+}\gamma{=}1$. RMSD is global C$\alpha$ RMSD in \AA.

\paragraph{Inverse folding (structure$\!\to$sequence):}
\[
R_{\mathrm{inv}}
= 0.60\,\mathrm{AAR}
+ 0.35\,\mathrm{scTM}
+ 0.05\,B,
\]
where AAR is residue-wise recovery versus the native sequence, scTM is the TM-score between the target backbone and the fold predicted from the designed sequence, and $B$ is a small biophysical bonus (e.g., motif/chemistry checks).

\paragraph{Motif scaffolding (motif$\!\to$scaffold):}
Candidates that do \emph{not} exactly preserve the motif receive $R_{\mathrm{motif}}{=}0$. Otherwise,
\[
R_{\mathrm{motif}}
= 0.40\,\overline{\mathrm{pLDDT}}_{\mathrm{scaf}}
+ 0.30\,g(\mathrm{RMSD}_{\mathrm{motif}})
+ 0.30\,\mathrm{scTM}
+ 0.20\,\mathbf{1}\!\big[\mathrm{RMSD}_{\mathrm{motif}}{<}1 \wedge \mathrm{scTM}{>}0.8\big],
\]
where $\overline{\mathrm{pLDDT}}_{\mathrm{scaf}}$ is mean pLDDT on non-motif residues, and the motif-aligned C$\alpha$ RMSD is mapped via
\[
g(x)=
\begin{cases}
\max(0,\,1-x/2), & x<1,\\[2pt]
\max(0,\,0.2-x/10), & x\ge 1.
\end{cases}
\]
We use a broad inverted normalization for \emph{global} folding RMSD and a sharper piecewise map (plus a success bonus) for \emph{local} motif RMSD to prioritize precise interface preservation.

\subsection{Length-binned improvements.}\label{app:length_binned}
\begin{figure}[h]
  \centering
  \includegraphics[width=\linewidth]{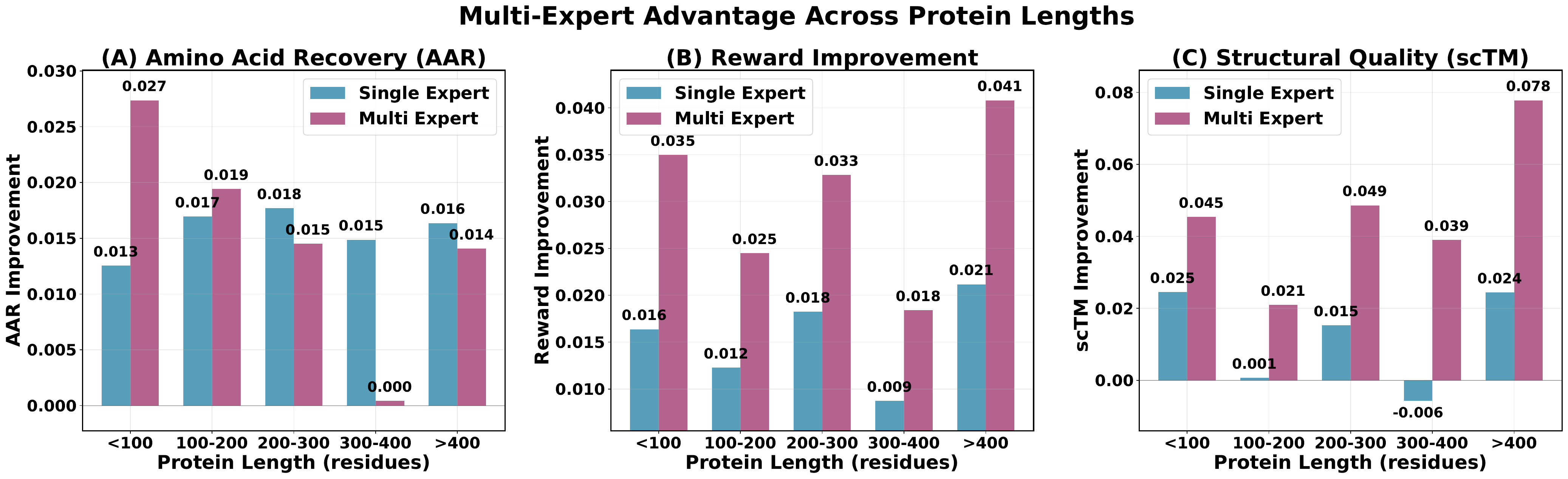}
  \caption{Length-binned improvements for inverse folding: mean $\Delta$AAR, $\Delta$ normalized reward, and $\Delta$scTM (Final$-$Baseline) for \emph{single-expert} vs.\ \emph{multi-expert} planners. Larger values indicate better recovery and structural consistency; the multi-expert planner yields consistent gains across bins, with larger margins on long proteins. }
  \label{fig:three_metric_comparison_app}
\end{figure}

Here we stratify inverse-folding gains by protein length to assess scalability. Figure~\ref{fig:three_metric_comparison_app} reports mean improvements (Final–Baseline) for AAR, normalized reward, and scTM across five bins: $<$100, 100–200, 200–300, 300–400, and $>$400 residues. The multi-expert planner outperforms the single-expert baseline in every bin, with especially pronounced margins on long proteins: scTM gains remain positive and grow with length (single-expert even turns slightly negative in the 300–400 bin), reward boosts are largest for $>$400, and AAR gains—while modest—are consistently higher for the ensemble (peak around 200–300). These trends indicate that expert diversity is most beneficial as sequence–structure complexity increases. \fix{AAR is a strict token-identity metric and thus less sensitive to small but structurally meaningful edits, whereas scTM/reward are smoother structural objectives that respond strongly to localized corrections. In addition, our critic emphasizes structural fidelity and the pLDDT-guided masking focuses changes on low-confidence regions, encouraging fold-improving (not necessarily native-identical) substitutions—producing larger scTM/reward deltas than AAR, especially for longer, more complex proteins.}

\subsection{Lead-optimization vs. De novo generation.}\label{app:denovo}
While our current implementation treats \fix{root} node as a complete sequence—effectively a lead-optimization style search—the framework naturally generalizes to fully de novo generation. In this setting, the root node could be a fully masked sequence, with subsequent nodes representing partially unmasked intermediates obtained through pLDDT-guided masking. Evaluation would then involve a simulation stage for reward estimation at each partial completion. Conceptually, this is equivalent to our guided unmasking scheme, differing only in granularity: entropy and disagreement are currently assessed at the level of entire candidate sequences, but under de novo generation they would be localized to the partially unmasked subsequence. As a result, performance is expected to be qualitatively similar, with the search dynamics adapting seamlessly from complete-sequence optimization to progressive de novo exploration.

\subsection{Related works.}\label{app:related}

\paragraph{Protein and \fix{Molecular} Diffusion Models.} 

Recent advances apply diffusion models to protein design across both structure and sequence domains. RFdiffusion \citep{Watson2023DeNovo} introduced a guided 3D diffusion model for protein backbones. On the sequence side, DPLM \citep{wang2024diffusionlanguagemodelsversatile} and its extension DPLM-2 \citep{wang2024dplm2multimodaldiffusionprotein} demonstrated that discrete diffusion pretraining on large-scale protein data enables high-quality sequence generation and versatile conditioning (e.g., inverse folding, property steering). Most recently, DiffGui \citep{hu2025diffgui} proposed a target-aware, E(3)-equivariant diffusion framework for joint atom and bond generation, incorporating multi-property guidance (e.g., binding affinity, drug-likeness) to generate realistic, high-affinity 3D molecules. However, these diffusion approaches generally rely on one-shot or open-loop decoding, making it difficult to incorporate adaptive search or revision during generation—highlighting the need for planning methods such as MCTS.

\paragraph{Monte Carlo Tree Search in Generative Modeling.}

Monte Carlo Tree Search (MCTS) has improved generative decoding in tasks requiring complex reasoning or program synthesis. In code generation, AlphaCode \citep{li2022competition} used a brute-force generate-and-test method by sampling large program sets and selecting those that passed tests. More structured methods followed: RethinkMCTS \citep{Li2024RethinkMCTSRE} performs MCTS over LLM reasoning steps, using execution feedback to refine intermediate thoughts. Tree-of-Thoughts \citep{yao2023tree} frames reasoning as a tree of self-evaluated partial solutions expanded via classical search (DFS/BFS), boosting performance on puzzles and planning. PG-TD \citep{zhang2023planning} simulates lookahead in code generation by executing candidate programs during decoding. 
ERP~\citep{Liu2024-ERP} achieves state-of-the-art performance by leveraging MCTS with a GPT model, with applications in both drug discovery and code generation.
Across domains, MCTS-style planning consistently enhances generation quality and reliability. Yet, most of these approaches are tied to autoregressive GPT-style models, which remain token-by-token and struggle with long-range dependencies—suggesting that discrete diffusion could provide a more flexible backbone for planning.

\paragraph{Discrete Diffusion for Planning and Generation.}

While originally developed for continuous domains, diffusion models have been successfully adapted to discrete sequence generation and planning. D3PM \citep{Austin2021StructuredDD} introduced absorbing-state (masking) noise for discrete diffusion, achieving competitive text generation. Follow-ups  diffusion language models like MD4 \citep{shi2024simplified} further closed the performance gap with autoregressive models. In planning, Diffuser \citep{janner2022diffuser} modeled offline RL as trajectory denoising, generating full state-action paths guided by value functions. To enable adaptive inference, Monte Carlo Tree Diffusion (MCTD) \citep{Baek2025MonteCD} reimagines diffusion as tree search, branching multiple denoising outcomes per step and scoring partial trajectories via heuristics. In protein design, ProtInvTree \citep{liu2025protinvtreedeliberateproteininverse} similarly integrates reward-guided search with jumpy denoising for efficient inverse folding. More recently, PepTune \citep{Tang2024PepTune} proposes multi-objective discrete diffusion for therapeutic peptide SMILES, pairing masked diffusion with an MCTS-style guidance strategy to negotiate tradeoffs among binding, solubility, and membrane permeability objectives. 
Still,  existing methods optimize toward a single diffusion generator, which limits their ability to effective search space exploration in biological and chemical design.

\paragraph{Multi-Experts Guided Generation.}
{Multi-expert systems leverage several generative models, or 'experts,' in tandem, allowing each to contribute its strengths while combining their state-wise expertise.
This idea has been explored in reinforcement learning~\citep{liu2023blending,liu2023active} and bandit~\citep{liu2024contextual} settings, where algorithms learn from multiple expert policies or oracles to improve decision-making. In generative modeling, mixture-of-experts strategies similarly combine specialized models to better cover complex data distributions. For example, recent diffusion-based image generators use multiple specialized denoising models across different noise levels \citep{fang2024remix} or even heterogeneous model architectures \citep{lee2024multi} to boost output quality. However, these multi-expert approaches remain largely static and one-shot during inference, without an adaptive search process. To our knowledge, no prior diffusion model integrates multiple experts into a planning loop for generative design. By incorporating a team of expert generators within an MCTS-driven diffusion framework, our work enables dynamic collaboration among experts to satisfy multiple design criteria and explore a broader solution space—an innovative combination that extends diffusion-based generation to more complex, multi-objective scenarios. } \fix{Complementary to search-based methods, DRaFT and ReFT-style approaches optimize generation via training-time reward alignment or parameter-efficient finetuning \citep{Clarketal2023, Luongetal2024}; in contrast, our framework is training-free and operates at test time by planning over diffusion rollouts.}

\subsection{Case study on CAMEO data and motif scaffolding.}

\begin{figure}[t]
  \centering
  \includegraphics[width=0.92\linewidth]{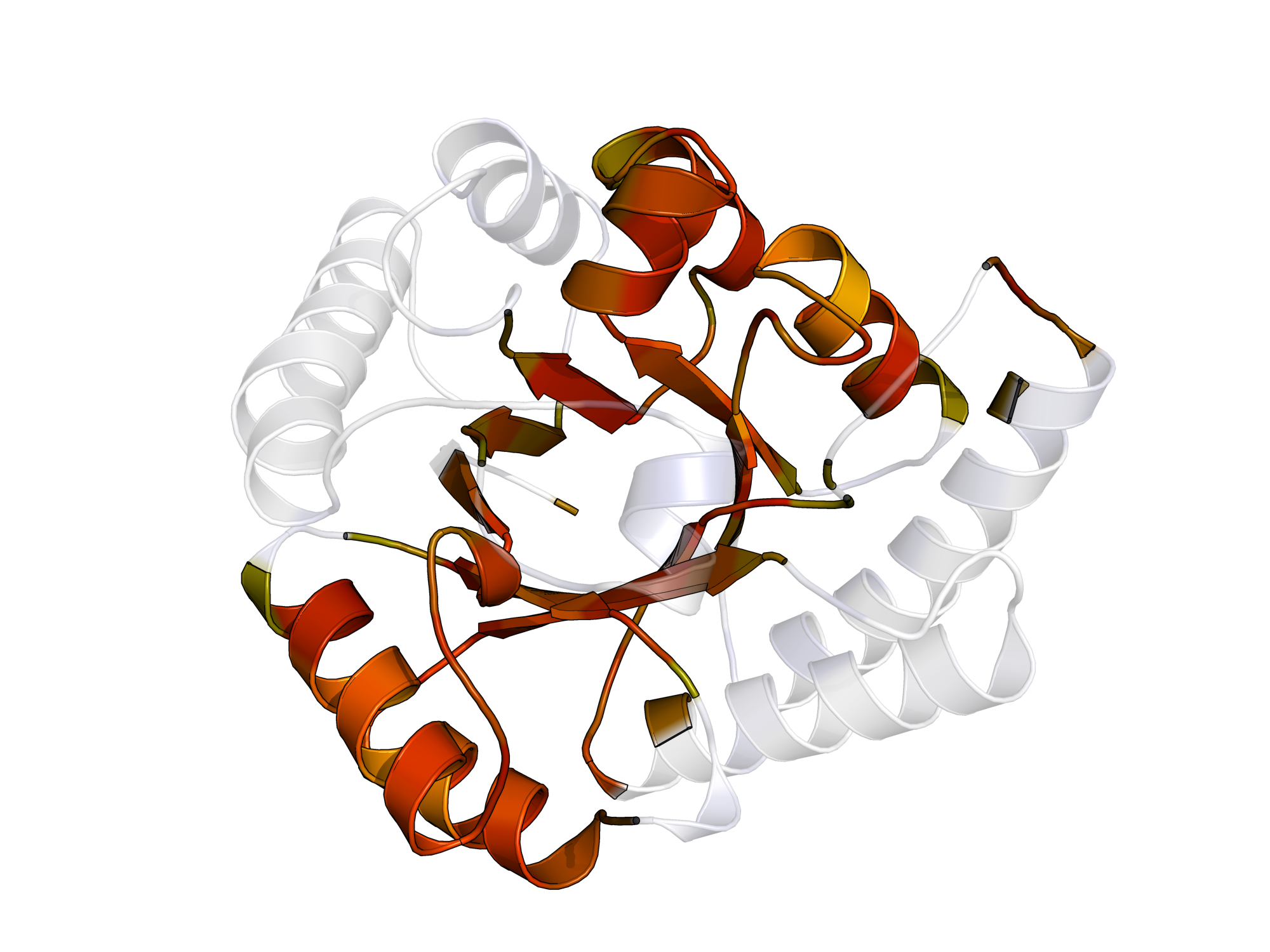}
  \caption{\textbf{CAMEO 7dz2\_C refinement.} 
  Light gray: baseline lead. 
  Colored overlay: \algname edits where improvement is achieved, shaded by closeness to the native backbone (red = very close; yellow = closer but not as close). 
  Gray segments indicate no change from the lead. 
  The predominance of red/yellow across many regions visualizes how \algname moves the lead toward the native geometry via pLDDT-aware masking and multi-expert selection.}
  \label{fig:cameo-7dz2c-cartoon}
\end{figure}

Figure~\ref{fig:cameo-7dz2c-cartoon} shows how \algname refines a CAMEO target (7dz2\_C). 
The light-gray ribbon is the \emph{baseline lead} (one-shot structure). 
Over it, the \algname design is drawn only where an improvement over the lead is made and is colored by per-residue closeness to the native backbone after alignment: \emph{red = very close}, \emph{yellow = closer but not as close}. 
(Regions without improvement are left in gray and reflect the baseline.) 
We observe broad reddening across helices and connecting loops, indicating tighter agreement with the target while preserving segments the lead already modeled well. 
This aligns with our mechanism: pLDDT-aware masking concentrates edits on low-confidence sites, and PH-UCT-ME promotes diverse, high-value proposals from multiple experts, yielding higher fold consistency without over-editing stable regions.

\begin{figure}[t]
  \centering
  \includegraphics[width=0.92\linewidth]{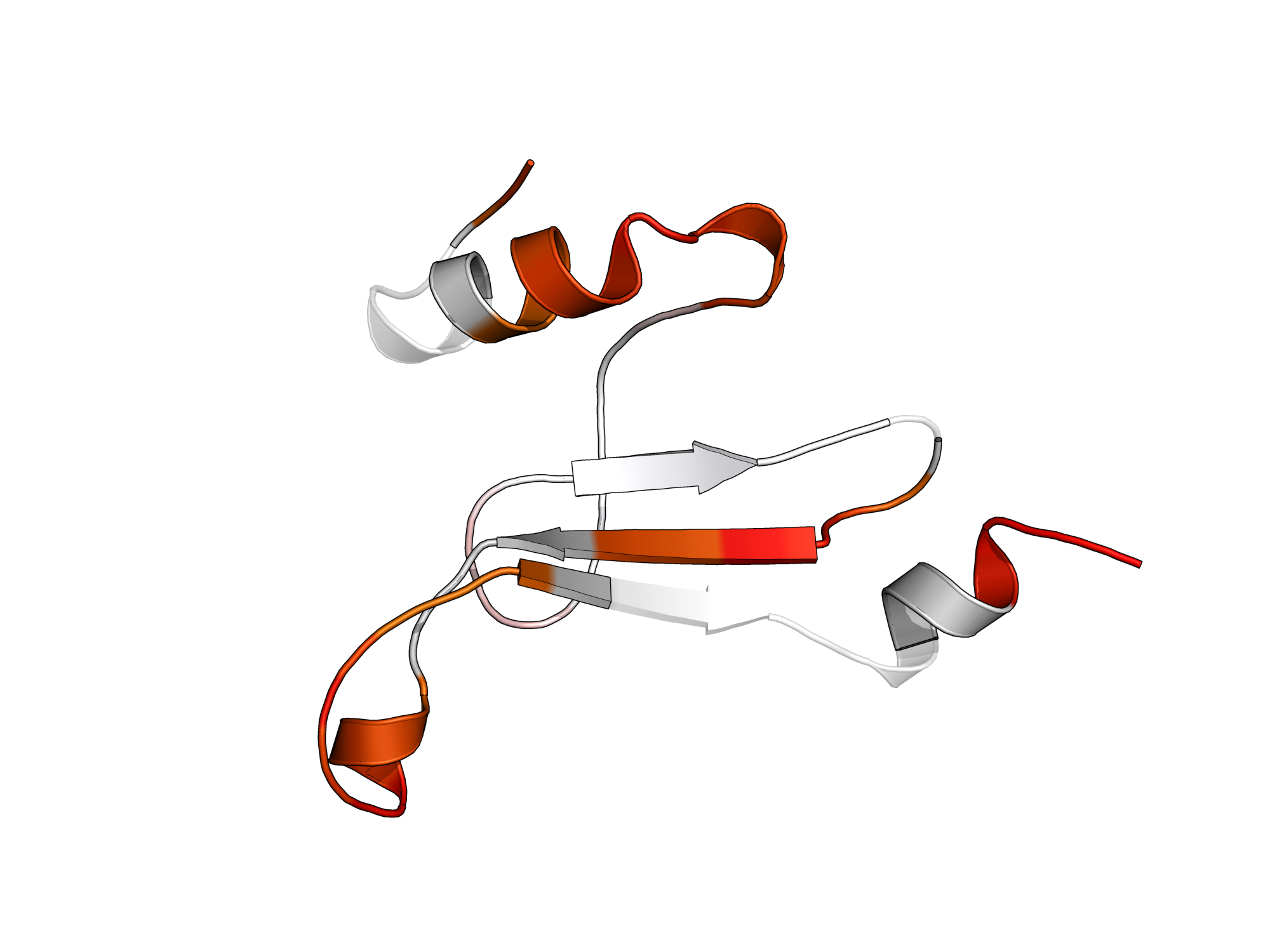}
  \caption{\textbf{Motif scaffolding (PDB 7mrx).} 
  Light/white segment: fixed input motif; gray ribbon: baseline scaffold (lead). 
  Colored overlay: \algname’s improved scaffold residues (red = very close to native; yellow = closer), shown only where the scaffold improves relative to the lead after motif-aligned superposition. 
  The localized reddening near the motif indicates targeted refinement of the interface without over-editing stable regions.}
  \label{fig:motif-7mrx}
\end{figure}

Figure~\ref{fig:motif-7mrx} visualizes how \algname refines the \emph{scaffold} around a fixed functional motif. 
The light, semi-transparent segment indicates the input motif (hard-frozen throughout generation). 
The surrounding gray ribbon is the baseline scaffold (lead), while the colored overlay shows \algname edits \emph{only where improvement over the lead is achieved}, shaded by closeness to the native structure after motif-aligned superposition (\emph{red = very close}, \emph{yellow = closer but not as close}). 
We observe concentrated reddening along the \(\beta\)-strand and adjacent loop, indicating better register and loop geometry at the motif–scaffold interface, while unchanged regions remain gray—highlighting that \algname preserves already adequate geometry and focuses edits where they matter.

\subsection{Additional experiments.}\label{app:additionalexperiments}

\subsubsection{Random-lead evaluation (model-agnosticity).}
\fix{To complement the fixed-lead analysis, we report inverse-folding results when the \emph{baseline} is a fresh one-shot sample per target (``random lead'') from DPLM-2 (150M), introducing stochastic variation across runs (see Table~\ref{tab:inverse_folding_results}). Despite this added baseline noise, \algname consistently attains the strongest improvements in normalized reward and scTM, while maintaining (CAMEO) or improving (PDB) AAR relative to single-expert controls and random routing. Taken together with the fixed-lead results in the main text, these outcomes demonstrate \textbf{model agnosticity}: the planner reliably refines \emph{any} provided lead—whether fixed or randomly resampled—under a shared critic. Minor differences in $\Delta$ magnitudes reflect baseline variability rather than changes to the optimization policy.}

\begin{table*}[t]
\centering
\resizebox{0.9\textwidth}{!}{%
\begin{tabular}{lccc|ccc|ccc}
\toprule
\multirow{2}{*}{\makecell[l]{Variant}} & \multicolumn{3}{c|}{AAR $\uparrow$} & \multicolumn{3}{c|}{Normalized Reward $\uparrow$} & \multicolumn{3}{c}{scTM $\uparrow$} \\
\cmidrule(lr){2-4} \cmidrule(lr){5-7} \cmidrule(lr){8-10}
 & Baseline & Final & $\Delta$ & Baseline & Final & $\Delta$ & Baseline & Final & $\Delta$ \\
\midrule
\multicolumn{10}{c}{\textbf{CAMEO Benchmark}} \\
\midrule
MCTD-0 (Random) 
& $0.446 \pm 0.011$ & $0.387 \pm 0.009$ & $-0.058 \pm 0.004$ 
& $0.417 \pm 0.009$ & $0.416 \pm 0.008$ & $-0.001 \pm 0.003$ 
& $0.378 \pm 0.013$ & $0.440 \pm 0.014$ & $+0.062 \pm 0.009$ \\
Single-Expert (150M) 
& $0.441 \pm 0.009$ & $0.441 \pm 0.009$ & $+0.000 \pm 0.000$ 
& $0.289 \pm 0.008$ & $0.289 \pm 0.008$ & $+0.000 \pm 0.000$ 
& $0.371 \pm 0.015$ & $0.371 \pm 0.015$ & $+0.000 \pm 0.000$ \\
Single-Expert (650M) 
& $0.443 \pm 0.009$ & $0.458 \pm 0.009$ & $+0.015 \pm 0.002$
& $0.435 \pm 0.011$ & $0.430 \pm 0.010$ & $-0.005 \pm 0.004$
& $0.371 \pm 0.015$ & $0.443 \pm 0.015$ & $+0.072 \pm 0.008$ \\
Single-Expert (ProteinMPNN) 
& $0.448 \pm 0.013$ & $0.392 \pm 0.012$ & $-0.056 \pm 0.005$
& $0.430 \pm 0.011$ & $0.422 \pm 0.010$ & $-0.007 \pm 0.004$
& $0.366 \pm 0.016$ & $0.448 \pm 0.015$ & $+0.082 \pm 0.009$ \\
\textbf{\algname (Multi-Expert)} 
& $\mathbf{0.452 \pm 0.013}$ & $\mathbf{0.445 \pm 0.012}$ & $\mathbf{-0.007 \pm 0.005}$
& $\mathbf{0.432 \pm 0.012}$ & $\mathbf{0.474 \pm 0.011}$ & $\mathbf{+0.041 \pm 0.006}$
& $\mathbf{0.342 \pm 0.018}$ & $\mathbf{0.445 \pm 0.018}$ & $\mathbf{+0.103 \pm 0.013}$ \\
\midrule
\multicolumn{10}{c}{\textbf{PDB Benchmark}} \\
\midrule
MCTD-0 (Random)                 
& $0.434 \pm 0.003$ & $0.401 \pm 0.003$ & $-0.033 \pm 0.002$ 
& $0.213 \pm 0.002$ & $0.200 \pm 0.002$ & $-0.013 \pm 0.001$ 
& $0.355 \pm 0.003$ & $0.376 \pm 0.003$ & $+0.021 \pm 0.002$ \\
Single-Expert (150M)            
& $0.460 \pm 0.006$ & $0.460 \pm 0.006$ & $+0.000 \pm 0.000$ 
& $0.290 \pm 0.006$ & $0.290 \pm 0.006$ & $+0.000 \pm 0.000$ 
& $0.370 \pm 0.007$ & $0.370 \pm 0.007$ & $+0.000 \pm 0.000$ \\
Single-Expert (650M)            
& $0.508 \pm 0.007$ & $0.541 \pm 0.007$ & $+0.033 \pm 0.003$ 
& $0.348 \pm 0.007$ & $0.382 \pm 0.006$ & $+0.034 \pm 0.003$ 
& $0.392 \pm 0.008$ & $0.426 \pm 0.008$ & $+0.034 \pm 0.003$ \\
Single-Expert (ProteinMPNN) 
& $0.451 \pm 0.016$ & $0.398 \pm 0.016$ & $-0.053 \pm 0.006$
& $0.330 \pm 0.013$ & $0.318 \pm 0.012$ & $-0.012 \pm 0.004$
& $0.384 \pm 0.009$ & $0.444 \pm 0.009$ & $+0.060 \pm 0.005$ \\
\textbf{\algname (Multi-Expert)}
& $\mathbf{0.504 \pm 0.006}$ & $\mathbf{0.578 \pm 0.006}$ & $\mathbf{+0.074 \pm 0.004}$ & $\mathbf{0.353 \pm 0.007}$ & $\mathbf{0.427 \pm 0.007}$ & $\mathbf{+0.074 \pm 0.004}$ & $\mathbf{0.383 \pm 0.007}$ & $\mathbf{0.448 \pm 0.008}$ & $\mathbf{+0.065 \pm 0.004}$ \\
\bottomrule
\end{tabular}}
\caption{\textbf{Inverse folding results on CAMEO and PDB.} 
{Baseline sequences are generated by one-shot DPLM-2 (150M)} , which serves as the initialization for our lead-optimization procedure. 
\emph{Baselines} may vary slightly across runs (see baseline-invariant results in \appref{app:additionalexperiments}), while {\emph{Final} denotes the sequence after \algname refinement. }
Metrics reported include amino acid recovery (AAR), normalized reward, and scTM, with {$\Delta$ indicating improvements over baseline.} 
 }
\vspace{-0.5cm}
\label{tab:inverse_folding_results}
\end{table*}

\subsubsection{One-shot expert baselines (no refinement)}

\paragraph{Inverse folding (one-shot experts).} 
Table~\ref{tab:oneshot_inverse_folding_experts_sctm_composite} reports one-shot inverse-folding quality for each \emph{expert} generator without any refinement. DPLM-2 scales favorably from 150M to 650M across AAR, scTM, and the composite metric; ProteinMPNN trades higher AAR for stronger biophysical plausibility (reflected in the composite). For a fair comparison to these single-shot experts, we also include a \emph{de novo} variant of \algname (MCTD-ME) that samples a single sequence per target using the same stochastic rollout settings. Note this setting differs from our lead-optimization regime used elsewhere; results here should therefore not be conflated with tables that evaluate refinement starting from a fixed lead.

\begin{table*}[h]
\centering
\small
\setlength{\tabcolsep}{7pt}
\renewcommand{\arraystretch}{1.12}
\resizebox{\textwidth}{!}{%
\begin{tabular}{l c c c}
\toprule
\textbf{Method} &
\textbf{AAR} (mean $\pm$ std) \ \textit{[min--max]} &
\textbf{scTM} (mean $\pm$ std) \ \textit{[min--max]} &
\textbf{Composite} (mean $\pm$ std) \ \textit{[min--max]} \\
\midrule
DPLM-2 150M (stoch) & $0.370 \pm 0.114$ \ \textit{[0.031--0.799]} & $0.321 \pm 0.061$ \ \textit{[0.087--0.490]} & $0.381 \pm 0.080$ \ \textit{[0.136--0.641]} \\
DPLM-2 650M (stoch) & $0.403 \pm 0.128$ \ \textit{[0.051--0.824]} & $0.326 \pm 0.060$ \ \textit{[0.093--0.501]} & $0.402 \pm 0.090$ \ \textit{[0.167--0.658]} \\
ProteinMPNN (stoch) & $0.242 \pm 0.064$ \ \textit{[0.029--0.360]} & $0.309 \pm 0.063$ \ \textit{[0.118--0.498]} & $0.300 \pm 0.054$ \ \textit{[0.107--0.387]} \\
\textbf{\algname} & $0.432 \pm 0.119$ \ \textit{[0.082--0.827]} & $0.329 \pm 0.056$ \ \textit{[0.124--0.497]} & $0.420 \pm 0.081$ \ \textit{[0.200--0.659]} \\
\bottomrule
\end{tabular}
}
\caption{\textbf{Inverse folding (CAMEO, $n{=}183$) with stochastic one-shot baselines plus \algname de novo.}
Each expert baseline emits a single sequence per target \emph{without refinement}, using \emph{stochastic sampling configured to mirror MCTS rollouts}: {\tt --temperature=1.0}, {\tt --unmasking\_strategy=stochastic1.0}, {\tt --sampling\_strategy=annealing@2.2:1.0}. 
We report amino-acid recovery (AAR), self-consistency TM-score (scTM), and the composite reward as mean $\pm$ std across targets, with min--max in brackets. 
\textbf{\algname (MCTS)} is shown for reference (refined result under planning).}
\label{tab:oneshot_inverse_folding_experts_sctm_composite}
\end{table*}

\paragraph{Forward folding (one-shot structures).} As a complementary check of generative fidelity, Table~\ref{tab:oneshot_forward_folding_dplm} evaluates one-shot forward folding (single prediction per sequence; no refinement). The 650M DPLM-2 exceeds the 150M across TM-score/RMSD under the same stochastic rollout settings, and \algname (de novo) is included for parity. The count is $n{=}163$ (not $183$) because the evaluator filters by length (default $60 \leq L \leq 512$), so out-of-range chains are skipped. As above, these \emph{de novo} results should not be conflated with lead-optimization tables.

\begin{table*}[h]
\centering
\footnotesize
\setlength{\tabcolsep}{6pt}
\renewcommand{\arraystretch}{1.1}
\resizebox{\textwidth}{!}{%
\begin{tabular}{@{}l c c c@{}}
\toprule
\textbf{Method} &
\textbf{TM-score to GT} (mean $\pm$ std) \ \textit{[min--max]} &
\textbf{BB RMSD to GT} (mean $\pm$ std) \ \textit{[min--max]} &
\textbf{Composite Reward} (mean $\pm$ std) \ \textit{[min--max]} \\
\midrule
DPLM-2\_150M\textsubscript{stoch} &
$0.656 \pm 0.168$ \ \textit{[0.296--0.932]} &
$11.015 \pm 6.410$ \ \textit{[2.228--27.337]} &
$0.333 \pm 0.136$ \ \textit{[0.122--0.596]} \\
DPLM-2\_650M\textsubscript{stoch} &
$0.724 \pm 0.160$ \ \textit{[0.299--0.937]} &
$8.400 \pm 5.819$ \ \textit{[1.852--30.628]}  &
$0.395 \pm 0.137$ \ \textit{[0.123--0.618]} \\
DPLM-2\_3B\textsubscript{stoch}   &
$0.345 \pm 0.129$ \ \textit{[0.174--0.873]} &
$40.777 \pm 11.735$ \ \textit{[7.062--68.432]} &
$0.142 \pm 0.054$ \ \textit{[0.073--0.371]} \\
\textbf{\algname{} (MCTS\_ME)}     &
$\mathbf{0.732 \pm 0.140}$ \ \textit{[0.325--0.934]} &
$\mathbf{8.331 \pm 5.570}$ \ \textit{[2.209--29.407]} &
$\mathbf{0.399 \pm 0.126}$ \ \textit{[0.133--0.631]} \\
\bottomrule
\end{tabular}%
}
\caption{\textbf{One-shot forward folding (CAMEO subset, $n{=}163$).}
Each model predicts a single structure per target from sequence (no refinement). We report backbone TM-score and backbone RMSD to ground truth, plus a composite reward (mean $\pm$ std across targets; min--max in brackets).
\emph{Sampling configuration (for stochastic rows)}: \texttt{--task folding \ \ --temperature 1.0 \ \  \ \ --sampling\_strategy "annealing@2.2:1.0" \ \ --max\_iter 150}. \algname{} (MCTS\_ME) uses the same rollout temperature/unmasking/sampling schedule, with planning via PH-UCT.}
\label{tab:oneshot_forward_folding_dplm}
\end{table*}

\paragraph{Motif scaffolding (one-shot scaffolds).}
Table~\ref{tab:motif_baseline_real_models_dplm650m} reports per-motif one-shot results for \textsc{DPLM-2} 650M on the curated motif set of \citep{Yim2024improved}.
Across 24 motifs (100 samples each), 17 motifs admit at least one feasible scaffold under our strict success criterion (377/2400 total successes).
We report per-motif success rate together with motif RMSD, scTM, and pLDDT, establishing a transparent baseline and motivating our refinement/planning evaluations on the standard filtered set (motifs with $\ge$1 success).

\subsubsection{Evaluation metrics for motif-scaffolding baselines.}
\label{app:motif_metrics}
\textbf{scTM (structure-based TM-score).} For each generated sequence we predict a backbone with ESMFold, then compute scTM by aligning this predicted structure to the corresponding reference PDB backbone using TM-align. We report the TM-score normalized by the reference chain length so that scores are comparable across proteins of different lengths.

\textbf{Motif RMSD (motif fidelity).} We compute motif RMSD between the motif region in the ESMFold-predicted structure and the motif region in the generated backbone (e.g., from RFDiffusion/ProteinA/FoldFlow). We select C$\alpha$ atoms and compute RMSD after optimal rigid-body superposition (centering and rotation).

\textbf{Success criterion.} A sample is counted as successful if it satisfies both scTM $> 0.8$ and motif RMSD $< 1\,$\AA.

\begin{table}[t]
\centering
\scriptsize
\setlength{\tabcolsep}{3.2pt}
\renewcommand{\arraystretch}{1.05}
\resizebox{\linewidth}{!}{%
\begin{tabular}{lcccccccccccccccc}
\toprule
\multirow{2}{*}{\textbf{Motif}} &
\multicolumn{4}{c}{\textbf{DPLM-2 650M}} &
\multicolumn{4}{c}{\textbf{RFDiffusion}} &
\multicolumn{4}{c}{\textbf{ProteinA}} &
\multicolumn{4}{c}{\textbf{FoldFlow}} \\
\cmidrule(lr){2-5} \cmidrule(lr){6-9} \cmidrule(lr){10-13} \cmidrule(lr){14-17}
& \textbf{Succ.} & \textbf{RMSD}$\downarrow$ & \textbf{scTM}$\uparrow$ & \textbf{pLDDT}$\uparrow$ 
& \textbf{Succ.} & \textbf{RMSD}$\downarrow$ & \textbf{scTM}$\uparrow$ & \textbf{pLDDT}$\uparrow$ 
& \textbf{Succ.} & \textbf{RMSD}$\downarrow$ & \textbf{scTM}$\uparrow$ & \textbf{pLDDT}$\uparrow$ 
& \textbf{Succ.} & \textbf{RMSD}$\downarrow$ & \textbf{scTM}$\uparrow$ & \textbf{pLDDT}$\uparrow$ \\
\midrule
1BCF        &  1/100 & 8.424 & 0.807 & 69.5 & 0/100 & 5.714 & 0.316 & 77.1 & 0/100 & 9.427 & 0.260 & -- & 0/100 & 7.471 & 0.194 & -- \\
1PRW        & 65/100 & 0.990 & 0.837 & 83.3 & 0/100 & 6.281 & 0.743 & 90.2 & 0/100 & 9.905 & 0.181 & -- & 0/100 & 7.607 & 0.164 & -- \\
1QJG        &  0/100 & 5.029 & 0.688 & 78.9 & 0/100 & 5.216 & 0.406 & 81.9 & 0/100 & 5.033 & 0.142 & -- & 0/100 & 4.700 & 0.164 & -- \\
1YCR        & 10/100 & 0.909 & 0.679 & 67.1 & 0/100 & 1.695 & 0.151 & 78.6 & 94/100 & 0.342 & 0.914 & -- & 1/100 & 2.744 & 0.364 & -- \\
2KL8        & 54/100 & 1.007 & 0.838 & 72.9 & 0/100 & 5.374 & 0.645 & 77.4 & 0/100 & 5.453 & 0.877 & -- & 0/100 & 10.127 & 0.106 & -- \\
3IXT        & 33/100 & 0.683 & 0.720 & 72.8 & 0/100 & 4.776 & 0.085 & 68.1 & 100/100 & 0.159 & 0.970 & -- & 0/100 & 7.965 & 0.128 & -- \\
4JHW        &  0/100 & 9.540 & 0.672 & 61.7 & 0/100 & 5.770 & 0.171 & 70.3 & 0/100 & 12.654 & 0.126 & -- & 0/100 & 12.927 & 0.138 & -- \\
4ZYP        &  1/100 & 2.701 & 0.630 & 67.6 & 0/100 & 3.682 & 0.220 & 70.5 & 0/100 & 4.930 & 0.175 & -- & 0/100 & 6.035 & 0.135 & -- \\
5IUS        &  0/100 & 6.403 & 0.569 & 56.3 & 0/100 & 8.989 & 0.393 & 63.0 & 0/100 & 10.629 & 0.188 & -- & 0/100 & 9.320 & 0.139 & -- \\
5TPN        &  0/100 & 2.578 & 0.710 & 61.0 & 33/100 & 1.506 & 0.636 & 76.3 & 99/100 & 0.436 & 0.892 & -- & 0/100 & 5.147 & 0.210 & -- \\
5TRV\_long  &  1/100 & 3.256 & 0.696 & 73.8 & 21/100 & 2.285 & 0.545 & 79.8 & 96/100 & 0.491 & 0.951 & -- & 0/100 & 6.309 & 0.277 & -- \\
5TRV\_med   &  1/100 & 2.416 & 0.763 & 74.8 & 2/100 & 3.572 & 0.281 & 76.7 & 89/100 & 0.524 & 0.949 & -- & 0/100 & 6.799 & 0.272 & -- \\
5TRV\_short &  0/100 & 3.550 & 0.696 & 68.8 & 0/100 & 5.579 & 0.077 & 81.1 & 83/100 & 0.509 & 0.950 & -- & 0/100 & 8.562 & 0.186 & -- \\
5WN9        &  0/100 & 4.989 & 0.559 & 61.3 & 0/100 & 5.207 & 0.171 & 66.1 & 100/100 & 0.156 & 0.934 & -- & 0/100 & 8.886 & 0.091 & -- \\
5YUI        &  0/100 & 6.493 & 0.623 & 69.9 & 0/100 & 3.397 & 0.481 & 70.2 & 0/100 & 6.596 & 0.187 & -- & 0/100 & 6.929 & 0.181 & -- \\
6E6R\_long  & 49/100 & 0.676 & 0.768 & 79.2 & 46/100 & 0.646 & 0.633 & 84.0 & 51/100 & 1.196 & 0.870 & -- & 0/100 & 3.378 & 0.471 & -- \\
6E6R\_med   & 42/100 & 0.653 & 0.760 & 79.8 & 40/100 & 0.707 & 0.504 & 83.1 & 68/100 & 0.873 & 0.876 & -- & 0/100 & 3.419 & 0.468 & -- \\
6E6R\_short &  4/100 & 0.641 & 0.670 & 76.6 & 19/100 & 0.935 & 0.349 & 81.9 & 66/100 & 0.915 & 0.875 & -- & 0/100 & 3.496 & 0.446 & -- \\
6EXZ\_long  & 35/100 & 1.117 & 0.772 & 74.8 & 56/100 & 0.938 & 0.639 & 84.7 & 100/100 & 0.180 & 0.924 & -- & 0/100 & 3.807 & 0.347 & -- \\
6EXZ\_med   & 35/100 & 1.106 & 0.771 & 74.7 & 44/100 & 1.354 & 0.511 & 81.3 & 100/100 & 0.172 & 0.928 & -- & 0/100 & 3.762 & 0.345 & -- \\
6EXZ\_short &  3/100 & 2.136 & 0.663 & 65.3 & 37/100 & 1.476 & 0.503 & 82.9 & 100/100 & 0.177 & 0.926 & -- & 0/100 & 4.182 & 0.329 & -- \\
7MRX\_128   & 13/100 & 2.484 & 0.735 & 69.8 & 19/100 & 2.204 & 0.476 & 72.9 & 100/100 & 0.165 & 0.943 & -- & 0/100 & 5.044 & 0.241 & -- \\
7MRX\_60    & 12/100 & 1.530 & 0.772 & 70.8 & 21/100 & 3.026 & 0.357 & 73.0 & 100/100 & 0.174 & 0.941 & -- & 0/100 & 5.305 & 0.208 & -- \\
7MRX\_85    & 18/100 & 1.986 & 0.809 & 71.4 & 17/100 & 2.666 & 0.380 & 72.2 & 100/100 & 0.174 & 0.942 & -- & 0/100 & 5.291 & 0.201 & -- \\
\bottomrule
\end{tabular}%
}
\caption{\textbf{One-shot motif scaffolding baselines (per-motif success and quality; Table~8).}
We evaluate each method as a one-shot scaffold generator on 24 curated motifs (following \citep{Yim2024improved}), with 100 independent generations per motif.
\textbf{Succ.} is the number of generations meeting our strict motif-preservation criterion.
We report motif-aligned RMSD (lower is better), self-consistency TM-score (scTM; higher is better), and mean pLDDT confidence (higher is better).
ProteinA and FoldFlow pLDDT values are unavailable in these runs and are reported as ``--''.}
\label{tab:motif_baseline_real_models_dplm650m}
\end{table}

\begin{table*}[h]
\centering
\small
\setlength{\tabcolsep}{7pt}
\renewcommand{\arraystretch}{1.15}
\begin{tabular}{lcccc}
\toprule
\textbf{Metric} & \textbf{DPLM-2 650M} & \textbf{RFDiffusion} & \textbf{ProteinA} & \textbf{FoldFlow} \\
\midrule
Motifs with $\ge 1$ success ($/24$) & 17/24 & 12/24 & 16/24 & 1/24 \\
Total successes ($/2400$) & 377/2400 (15.7\%) & 355/2400 (14.8\%) & 1446/2400 (60.2\%) & 1/2400 (0.0\%) \\
Mean motif RMSD $\downarrow$ & $2.971 \pm 3.278$ & $3.382 \pm 2.583$ & $2.969 \pm 3.984$ & $6.217 \pm 2.610$ \\
Mean scTM $\uparrow$ & $0.717 \pm 0.129$ & $0.403 \pm 0.362$ & $0.705 \pm 0.340$ & $0.242 \pm 0.148$ \\
Mean pLDDT $\uparrow$ & $70.9 \pm 11.1$ & $76.8 \pm 11.9$ & -- & -- \\
\bottomrule
\end{tabular}
\caption{\textbf{Motif scaffolding one-shot baselines (Table~8; summary over 24 motifs, 100 samples each).}
DPLM-2 650M numbers are from Table~\ref{tab:motif_baseline_real_models_dplm650m}.
RFDiffusion numbers follow the updated full-pipeline baseline (ProteinMPNN $\rightarrow$ ESMFold); mean RMSD is computed over motifs with defined RMSD.
ProteinA and FoldFlow pLDDT values are unavailable in these runs and are reported as ``--''.}
\label{tab:motif_baseline_real_models}
\end{table*}

ProteinA is substantially stronger than FoldFlow on this motif-region evaluation: it achieves a $60.2\%$ overall success rate (1446/2400) and succeeds on 16/24 motifs, whereas FoldFlow only has 1/2400 successes (1/24 motifs with any success). Consistent with this, ProteinA attains much higher mean scTM ($0.705$ vs. $0.242$) and lower mean motif RMSD ($2.969$ vs. $6.217$), indicating markedly better motif fidelity and global structural agreement under our strict criterion.

\emph{Compute note.} The baseline numbers above are from 100 independent one-shot samples per motif. In contrast, in our main refinement/planning experiments we typically run only 25 search iterations per motif/target; despite this much smaller budget, the planner already surfaces many successful scaffolds in practice, demonstrating that multi-step guided search can identify high-quality candidates efficiently.

\subsection{Empirical decay of the P$\mathcal{H}$-ME bonus under tight compute}
\label{app:ph_decay}

When test-time compute is scarce, we attenuate the P$\mathcal{H}$-ME bonus so that
selection increasingly resembles a plain UCT rule. Let
\[
\rho \;\triangleq\; \tfrac{t}{T}\in[0,1],\qquad
B(s,a) \;\triangleq\; w_{\mathrm{ent}}\,U_{\mathrm{ent}}(s,a) \;+\; w_{\mathrm{div}}\,U_{\mathrm{div}}(s,a),
\]
where $t$ is the global selection step, $T$ is the total budget, $U_{\mathrm{ent}},U_{\mathrm{div}}\in[0,1]$,
and the nonnegative weights are chosen so that $B(s,a)\in[0,1]$ for all $(s,a)$.
We apply a power decay to the bonus:
\[
B_{\mathrm{decay}}(s,a;\rho) \;=\; \bigl(B(s,a)\bigr)^{\,1-\rho}.
\]
The selection score is then
\[
\PHUCBME^{\mathrm{decay}}(s,a)
\;=\;
Q(s,a)\;+\;
c_p\,\frac{\sqrt{\log N(s)}}{1+N(s,a)}\;
\pi_{\mathrm{cons},\tau}(a\mid s)\;\bigl(B(s,a)\bigr)^{\,1-\rho}.
\]
For early search ($\rho\!\approx\!0$), $\bigl(B(s,a)\bigr)^{1-\rho}\!\approx\!B(s,a)$ and the full
P$\mathcal{H}$-ME bonus is active, promoting exploration via large uncertainty/disagreement
($U_{\mathrm{ent}}$) and novelty ($U_{\mathrm{div}}$). As $\rho\!\to\!1$, $\bigl(B(s,a)\bigr)^{1-\rho}\!\to\!1$,
and the rule collapses to the two-term variant (UCT scaled by the consensus prior), thereby
favoring exploitation through $\pi_{\mathrm{cons},\tau}(a\mid s)$. In the extreme low-budget regime,
fixing $\rho\!=\!1$ recovers the two-term score exactly (pure exploitation modulated by UCB and the
consensus prior).

\paragraph{Notes.}
(i) The decay acts only on the P$\mathcal{H}$-ME bonus; the factor
$\sqrt{\log N(s)}/(1+N(s,a))$ is unmodified, so the classical UCB explore–exploit trade-off remains intact.
(ii) A smoothly varying alternative is to decay the exponent, e.g.,
$1-\rho=\tfrac{1+\cos(\pi\rho)}{2}$, which behaved similarly in practice.
(iii) When a visit-driven proxy is preferred without altering UCB, one may replace $\rho$
by $\tilde\rho(s)=\min\!\bigl(1,\,N(s)/N_{\mathrm{cap}}\bigr)$ with a user-chosen cap
$N_{\mathrm{cap}}$; this is a heuristic surrogate for progress and preserves the same form.

\subsection{Experimental configuration and hyperparameters.}\label{app:exp_config}\label{app:setup}

\paragraph{Search setup and hyperparameters.}
\begin{table}[ht]
\centering
\small
\begin{tabular}{lcc}
\toprule
\textbf{Component} & \textbf{Symbol} & \textbf{Setting(s) used} \\
\midrule
Rollouts per expert & $k_{\text{roll}}$ & \textbf{3} (fixed for reported runs) \\
Children kept per expansion & $K$ & \textbf{3} (beam size per expansion) \\
Forward steps (tree depth) & $\text{max\_depth}$ & \textbf{5} (default; held fixed unless noted) \\
Exploration constant (\PHUCTME) & $c_p$ & \textbf{1.414} (UCB1 constant) \\
Candidates per expansion (random mode) & $N_{\text{cand}}$ & \textbf{6} (used only in random\_no\_expert) \\
Backup rule & --- & \textbf{max} (value backup along the path) \\
Number of experts (multi\_expert) & $E$ & \textbf{3} (unless otherwise noted) \\
Modes compared & --- & random\_no\_expert,\; single\_expert,\; multi\_expert \\
Single-expert ID in plots & --- & \textbf{single\_expert\_0} (for direct comparisons) \\
\midrule
Diffusion reverse steps & $\text{max\_iter}$ & \textbf{150} \\
Diffusion temperature & $T$ & \textbf{1.0} \\
Decoding strategy (all modes) & --- & deterministic unmasking,\; argmax sampling \\
\bottomrule
\end{tabular}
\caption{\textbf{Search setup and hyperparameters.} We keep decoding and evaluation identical across methods. Hyperparameters are fixed for the main comparisons; only the presence/number of experts differs by mode.}
\end{table}
For equitable assessment, we compare all variants under matched search and decoding budgets. Unless otherwise specified, every method uses the same DPLM-2 decoding configuration (deterministic unmasking with argmax sampling) and identical reward evaluation. In each table/figure, competing methods share the same rollout budget and selection rule; only the availability of diffusion \emph{experts} differs across modes (random\_no\_expert, single\_expert, multi\_expert). We conducted only light tuning around a small grid and then fixed a single setting for all reported comparisons.

\paragraph{Models, experts, and baselines (full details for \S5.1).}
We instantiate \algname with masked \emph{discrete} diffusion as the rollout engine and vary the \emph{proposal experts} by task. Sequence experts are pre-trained DPLM-2 models \citep{wang2024diffusionlanguagemodelsversatile} of different capacities (150M, 650M, 3B). For motif scaffolding we consider structure-aware experts (RFDiffusion \citep{Watson2023RFdiffusion}, ProteinA \citep{Geffner2025Proteina}, FoldFlow \cite{Bose2024FoldFlow}) both as single-expert baselines and as members of a multi-expert ensemble. All experts propose from the \emph{same} masked input; a single composite critic scores candidates. Baselines are task-specific: \textit{Folding} (seq $\!\to\!$ struct) uses one-shot ESMFold \citep{lin2024esmfold_science} on the native sequence (RMSD/TM vs.\ ground truth), while our planners propose structure-token edits via DPLM-2 conditioned on the sequence and compare directly to the reference backbone. \textit{Inverse folding} (struct $\!\to\!$ seq) uses one-shot DPLM-2 (150M) conditioned on backbone tokens; \algname performs masked amino-acid fills given the backbone. \textit{Motif scaffolding} (motif $\!\to\!$ scaffold) uses one-shot DPLM-2 (150M) co-generating sequence/structure with motif residues frozen; \algname edits scaffold positions (sequence and/or structure, depending on expert) while preserving the motif.
\fix{In this study, we frame evaluation as \emph{lead optimization}: starting from a one-shot initial \emph{lead} and refine it. }
We emphasize that our goal is \emph{not} to propose a new one-shot SOTA generator/backbone.
Instead, MCTD-ME is a \emph{test-time sampling/guidance layer} that improves a fixed initial lead via discrete masked edits under a fixed evaluation budget.
Accordingly, we focus on \emph{Final} and especially $\Delta$ (Final–Baseline) under equal compute; stronger backbones can be incorporated as additional experts within the same planning loop.

We compare {MCTD-0} (random fill; no experts), {MCTD-1} (single proposal expert; separate runs per expert), and {\algname} (multi-expert ensemble). \fix{We also include two multi-expert ablations on the CAMEO subset ($n{=}183$), keeping the \emph{initial leads}, critics, and wall-clock budget fixed: (i) \emph{Sampling (no MCTS)}—experts propose directly on the pLDDT-derived mask; we score with the composite reward and report the per-target best (no selection/backprop); and (ii) \emph{UCT only}—MCTS with 2 UCB selection \citep{kocsis2006bandit} using $\mathrm{UCB}=Q(s,a)+c_p\sqrt{\tfrac{\log N(s)}{N(s,a)}}$, omitting the uncertainty/novelty bonuses used in \PHUCTME. Sampling isolates the benefit of structure-aware masking and expert diversity but underperforms \algname on recovery-style objectives and shows higher variance; UCT recovers part of the planning benefit over Sampling yet remains below \algname due to reduced proposal diversity.} \fix{Due to limited space, we defer additional baselines and experiments to \appref{app:additionalexperiments}.}

\subsection{Runtime and efficiency.}
The enhanced performance comes at the cost of additional computation. \algname must evaluate multiple experts across many tree expansions, whereas a single diffusion model decoding is much faster. In our current implementation, generating one sequence for a CAMEO structure with \algname-E takes on the order of a few minutes (wall-clock) on a single GPU – roughly an order of magnitude slower than a single DPLM-2 pass. We emphasize that our focus here is on proof-of-concept effectiveness; optimization of the search procedure is left for future work. Techniques such as parallelizing rollouts, caching expert evaluations, or pruning low-reward branches more aggressively could greatly speed up \algname without sacrificing quality.

Experiments were run on a SLURM-managed cluster with one NVIDIA A100 GPU per job, four CPU cores and 32 GB RAM. 
All results were obtained with PyTorch mixed precision and identical inference code across modes; wall-time numbers in Table~\ref{tab:runtime} are measured end-to-end per target (including expert rollouts and critic scoring).

\begin{table}[t]
\centering
\setlength{\tabcolsep}{10pt}
\begin{tabular}{lrrr}
\toprule
\textbf{Mode} & \textbf{Min (s)} & \textbf{Mean (s)} & \textbf{Max (s)}  \\
\midrule
Random (no expert) & 89.6  & 645.3 & 5235.6 \\
Single-expert (650M) & 154.9 & 345.7 & 1024.8\\
Single-expert (150M) & 257.6 & 502.9 & 1791.6 \\
Single-expert (3B)   & 148.6 & 451.7 & 1052.0 \\
Multi-expert (3$\times$) & 314.0 & 800.2 & 2187.2 \\
\bottomrule
\end{tabular}
\caption{\textbf{Per-target wall-clock time} (seconds) aggregated over CAMEO2022 and PDB\_date runs. 
“Single-expert (650M/150M/3B)” correspond to our three DPLM-2 capacities (IDs 0/1/2). 
Times include all steps (masking, diffusion rollouts, scoring, and tree updates).}
\label{tab:runtime}
\end{table}

\subsection{Generality.}
Although our evaluation focused on the inverse folding problem, the \algname framework is inherently general and model-agnostic. Any diffusion-based generator can be plugged into the tree search, and any collection of expert predictors can shape the reward. This flexibility opens up many possibilities: for instance, \algname could be applied to de novo protein design by incorporating structural-validity scores and evolutionary metrics as critics, or extended to other domains such as molecule generation using pharmacophoric and ADMET property predictors. Our results on protein inverse folding demonstrate the promise of combining diffusion models with MCTS guided by multiple critics for multi-objective generative design. We believe this diffusion+MCTS approach offers a powerful new paradigm for generative modeling, and extending \algname to more tasks with richer reward functions is an exciting direction for future work.

\end{document}